\documentclass{article}
\usepackage{slnotation}
\usepackage[subrefformat=parens]{subcaption}
\usepackage{rotating}
\usepackage{amssymb}

\newcommand{\sr}[0]{\evalMetric_{\mathrm{SR}}}
\newcommand{\pr}[0]{\evalMetric_{\mathrm{PR}}}

\newcommand{\spl}[0]{\evalMetric_{\mathrm{SPL}}}

\newcommand{\return}[0]{R}
\newcommand{\breakpoint}[0]{\hat{\return}}

\newcommand{\targetReturn}[0]{\return^*}

\newcommand{\returnAtTime}{f}
\newcommand{\returnAtTimei}{\returnAtTime}
\newcommand{\returnAtTimej}{\returnAtTime'}
\newcommand{\timeToReturn}{g}
\newcommand{\timeToReturni}{\timeToReturn}
\newcommand{\timeToReturnj}{\timeToReturn'}

\newcommand{\lr}[0]{\mathrm{LR}}
\newcommand{\rpp}[0]{\mathrm{RPP}}
\newcommand{\ipp}[0]{\mathrm{IPP}}
 \PassOptionsToPackage{numbers, compress}{natbib}

\usepackage[hidelinks]{hyperref}
\usepackage[eandd, final]{neurips_2026}

\usepackage[utf8]{inputenc} %
\usepackage[T1]{fontenc}    %
\usepackage{hyperref}       %
\usepackage{url}            %
\usepackage{booktabs}       %
\usepackage{amsfonts}       %
\usepackage{nicefrac}       %
\usepackage{microtype}      %
\usepackage{xcolor}         %

\newif\ifdegradationplots
\degradationplotstrue
\title{Offline Preference-Based Trajectory Evaluation}

\author{%
Fernando Diaz\\
Carnegie Mellon University\\
Pittsburgh, PA\\
\texttt{diazf@acm.org}\\
}

\begin{document}

\maketitle

\begin{abstract}
    Offline evaluation of agentic systems often collapses trajectories into a binary terminal success value, discarding information about partial progress.  This, in turn, results in widespread ties that create statistical inefficiency by reducing effective sample size and weakening the ability to distinguish systems. We propose preference-based trajectory evaluation, an evaluation measure that compares pairs of trajectories directly by using temporal preferences between time-to-return profiles. We find that, across diverse agentic and interactive benchmarks, success-based metrics result in tied comparisons on roughly 75\% of instances, while trajectory-aware preferences reduce ties to roughly 35\%, improving discriminative power, ranking stability, and data efficiency. Our results suggest that benchmark saturation, often portrayed as the result of poor data collection or problem difficulty, may also be explained by the design of the evaluation measure. 

\end{abstract}
\section{Introduction}
\label{sec:introduction}
Motivated by the increasing cost of evaluating AI systems \cite{ghosh:eval-cost}, we 
study two desiderata when measuring system performance: sensitivity and data efficiency. \textit{Sensitivity} \cite{mandel:sensitivity}, in the context of AI evaluation, refers to a measure's ability to detect meaningful performance differences between systems under a fixed evaluation budget. Sensitivity becomes important as the performance of AI systems improves to a quality where even substantively different behaviors can result in small observed differences. \textit{Data efficiency} is related to statistical power \cite{cohen1988statistical} and refers to the number of evaluation samples required to reach reliable comparative assessment of systems. Although distinct, these two objectives are related since insensitive metrics require substantially larger evaluation sets to resolve the same system differences that more sensitive metrics can detect with fewer observations. %

\begin{figure}
    \centering
    \begin{subfigure}[t]{0.35\textwidth}\centering
        \includegraphics[width=\linewidth]{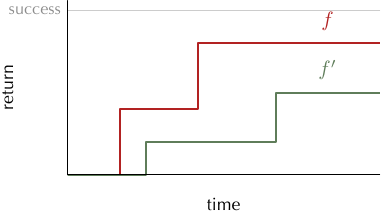}
        \caption{$\sr(\returnAtTimei)=\sr(\returnAtTimej)=0$}\label{fig:tied-failures}
    \end{subfigure}\hspace{1cm}
    \begin{subfigure}[t]{0.35\textwidth}\centering
        \includegraphics[width=\linewidth]{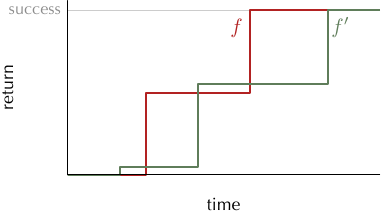}
        \caption{$\sr(\returnAtTimei)=\sr(\returnAtTimej)=1$}\label{fig:tied-successes}
    \end{subfigure}
    \caption{\textbf{Trajectory ties under success rate.} \subref{fig:tied-failures} Two unsuccessful trajectories can be distinguished by partial terminal returns.  \subref{fig:tied-successes} Two successful trajectories can be distinguished by accumulation of return over time.}
\end{figure}

Unfortunately, many contemporary evaluation approaches measure performance using binary success, a relatively insensitive and inefficient metric that answers the question, `did the system ever solve the task instance?' While  convenient, in agentic systems, basing a metric on binary success measurements poses two problems.  First, binary measurement often collapses partial solutions to 0, losing granular evaluation signals and conflating trajectories that may differ in progress toward a solution (Figure \ref{fig:tied-failures}).  Second, binary measurement collapses the performance accrued over multi-step trajectories into a single scalar value, comparing two trajectories using their terminal values instead of how performance develops over time (Figure \ref{fig:tied-successes}).  Combined, these two issues can compromise sensitive and efficient evaluation.  To understand how, we can look at the number of ties between systems when using success as a measure; a large number of ties degrades the effective evaluation set size.  In the benchmarks we study, an average of 75\% of instance-level comparisons are ties under success rate.  Even when comparing partial returns, the tie rate remains high at 50\%.  As systems become more performant, this inefficiency compounds because more trajectories are collapsed as indistinguishably (un)successful, prompting claims of benchmark saturation \cite{ott:saturation,vania:irt-saturation,kiela:dynabench}.  At the same time, the use of binary success metrics across machine learning and natural language processing conferences is increasing.  We found that the percentage of abstracts in papers published at the NeurIPS Datasets and Benchmarks track mentioning binary metrics rose from 5\% in 2022 to 18\% in 2025; at EMNLP, the fraction rose from 9\% in 2022 to 21\% in 2025 (details in Appendix \ref{app:conference-percentages}).  Together, these observations suggest that the current use of success rate is both inefficient and growing in adoption.  

While many existing approaches to address evaluation inefficiency focus on reducing the number of test instances while maintaining success rate as the metric, we approach insensitivity and inefficiency by interrogating the systematization of `performance' itself. Instead of collapsing each trajectory to a binary terminal value, we expand performance to capture how return progresses over time.  This allows us to compare trajectories by adopting the principle of temporal preference: given two systems achieving the same task progress, we prefer the system that reached it sooner.  
Inspired by recent results in information retrieval evaluation \cite{moffat:freedom,diaz:rpp}, we operationalize this principle in a family of measures that directly capture the preference without intermediary scalar metric computation.  
Importantly, our approach makes few assumptions beyond temporal preference and requires no additional hyperparameters, unlike methods based on temporal discounting.  

We assess our proposed methods across multiple benchmark families spanning both classic reinforcement learning environments and contemporary agentic tasks. Our results show that trajectory-aware preferences reduce tie rates from roughly 75\% to 35\% on average, recovering a substantial amount of signal. By preserving information, our methods lead to consistent improvements across standard measurement criteria: higher reliability, sensitivity, and data efficiency.  
More broadly, our results suggest that benchmark saturation can result from the information loss induced by metric definition. 
While the benefits of our approach may appear intuitive, current benchmarks adopt success rate, and the statistical consequences of this design choice have not been systematically studied.

\section{Background}
\label{sec:motivation}
Modern evaluations face two fundamental limitations. First, as models improve, many established benchmarks show signs of saturation and can no longer reliably distinguish high-performing systems \cite{ott:saturation,vania:irt-saturation,kiela:dynabench,akhtar:saturation}. Second, large-scale evaluation has become prohibitively expensive. \citet{ghosh:eval-cost} show that evaluating a single system on a modern benchmark can require several thousand dollars in inference costs alone. Taken together, these trends suggest that the benchmarks that are most expensive to run are often those least able to resolve meaningful differences between models.

As a result of these limitations, there have been increasing calls for more principled and rigorous evaluation \cite{olteanu:rigor}.  Approaches can be roughly divided into two categories.  The first category fixes the evaluation metric and develops methods to sample or weight instances to improve efficiency.  Methods include dynamic benchmarking \cite{ma:dynaboard,subramani:simba}, robust statistical practices \cite{agarwal:rleval,henderson:drl-that-matters,chouldechova:comparison-measurement}, active learning approaches \cite{kossen:active-testing,huang:active-testing,ashury-tahan-etal-2024-label,li:active-evaluation,mohankumar-khapra-2022-active}, and item response theory \cite{martinez-plumed:ai-evaluation-irt,vania:irt-saturation,truong:reliable-and-efficient-amortized-eval,rodriguez:example-importance,maia-polo:tiny-benchmarks,ndzomga:mid-difficulty}.  The second category focuses on the development of improved measurement instruments by adopting methods from measurement theory to design metrics \cite{wallach:measurement,chouldechova:comparison-measurement}, providing a theoretical framework to inspect the systematization of a concept (i.e., the relevant factors considered when measuring the concept) and its operationalization (i.e., how we detect and quantify the relevant factors).  In the case of success rate, `performance' may be systematized as `whether the agent completed the task' and operationalized as `whether the agent is in a pre-defined end state.'  Recent calls to consider system cost during evaluation \cite{kapoor:hal} can be interpreted as expanding the systematization of performance to include inference cost, which is then operationalized as `distance from the Pareto frontier of cost and task completion.'

\begin{table}
    \begin{center}
    {\small
    \begin{tabular}{cccccc}
        \toprule
        &\multicolumn{2}{c}{A}&\multicolumn{2}{c}{B}&\\
\cmidrule(lr){2-3} \cmidrule(lr){4-5}
        task & success & time & success & time & winner\\
        \hline
        1 & 1 & 2 & 1 & 1 & B\\
        2 & 1 & 11 & 1 & 10 & B\\
        3 & 1 & 11 & 1 & 10 & B\\
        4 & 1 & 11 & 1 & 10 & B\\
        5 & 1 & 1 & 0 & - & A\\
        \hline
        mean & 1 & 7.2 & 0.8 & 7.75 & -\\
        \bottomrule
    \end{tabular}
    }
    \end{center}
    
    \caption{\textbf{Aggregation reversal in decoupled success-rate and time-to-success evaluation.} Example success rate and conditional mean time-to-success for two models over five tasks.  The last column reflects which model has a faster time-to-success for the task.  Separately computing success rate and mean time-to-success erroneously suggests that A dominates B across both metrics when in fact B dominates A in 80\% of tasks.   }\label{tab:success-time-example}
\end{table}

Among dimensions that contribute to performance, time plays an important role for agentic systems.  Time efficiency has long been an important factor in system performance, changing the question from `can the system complete the task?' to `can the system complete the task in a reasonable amount of time?'  This is natural in the evaluation of agents since they complete tasks by interacting with the environment over multiple steps.  While many benchmarks calculate the average number of steps in addition to success rate, inspecting these metrics independently can conceal task instances where two systems succeed but with very different time efficiency, resulting in system order reversals  (Table \ref{tab:success-time-example}).  The temporal choice literature distinguishes between temporal preference---an ordering over outcomes with identical utility occurring at different times---and temporal discounting---a particular modulation of an outcome's utility based on when it occurs in time \cite{frederick:time-discounting}. In the context of comparing two trajectories, temporal preference would specify that, if both trajectories succeed, prefer the shorter trajectory.  Temporal discounting, by contrast, converts time into a scalar weight applied to performance, reducing the ordering of trajectories to comparing discounted utilities. 
In reinforcement learning, linear discounting (subtracting a constant penalty per step) and exponential discounting (rescaling future rewards multiplicatively by a constant factor) are often introduced for algorithmic convenience in optimization rather than as a principled evaluative criterion, which is often the undiscounted return \cite{barto-sutton:RL}. When time is considered, researchers often use power-law discounting, reflected in the `Success weighted by Path Length' metric  \cite{anderson:evaluation-of-navigation-agents}, which divides binary success by the ratio of trajectory length over the optimal path length. %
That said, in real world settings, temporal discount rates can vary by domain and are non-stationary \cite{frederick:time-discounting}, making these methods brittle since they assume a precise relationship between time and utility.  To avoid these issues, we focus on evaluation methods based on temporal preference, making fewer assumptions and avoiding additional hyperparameters.

Temporal preference is part of a broader class of approaches that shift from assigning scalar values to model outputs or behaviors (metric-based evaluation) to assigning signed values to pairs of model outputs (preference-based evaluation).  Although evaluation based on paired comparisons is an established method for variance reduction and improved data efficiency \cite{peyrard-etal-2021-better}, these methods have been largely absent from offline machine learning evaluation.  When paired comparisons arise, it is normally through online preference-based evaluation \cite{chiang:chatbot-arena,joachims:ranking-svm}, where explicit or implicit feedback from human users is used to assess which of two models' outputs is preferred.  When using online methods, comparing a new model requires collection of new data, which can be prohibitive during model development, due to experimentation speed and safety requirements. Our work can be seen as the offline counterpart of online arena-style preference-based evaluation.  As such, it inherits the benefits of offline evaluation, including counterfactual analysis (i.e., comparing more than two systems in the same context), safety, and speed.  

\section{Preference-Based Trajectory Evaluation}
\label{sec:metrics}
We are interested in broadening the systematization of performance beyond binary success to include richer trajectory information.
We represent a trajectory as a function $\slDef{\returnAtTime}{\slPInts}{[0,1]}$, where $\returnAtTime(t)$ is the normalized return at discrete time step $t$ and a return of $1$ indicates task success.  We consider domains where incremental rewards are non-negative and, as a result, $\returnAtTime(t)$ is nondecreasing in $t$. The \textit{time-to-return} is a function $\slDef{\timeToReturn}{[0,1]}{\slPInts\cup\{\infty\}}$, where $\timeToReturn(\return)$ is the first time at which the agent achieves a return of at least $\return$; if $\return$ is never reached, $\timeToReturn(\return) = \infty$.

An \textit{evaluation metric} is defined as $\evalMetric(\returnAtTime)\in[0, 1]$.  At an instance level, success rate (SR) systematizes performance as (binary) task completion, $\sr(\returnAtTime)=\slIndicator{\timeToReturn(1)<\infty}$; time and partial progress are both excluded.  Partial return (PR) systematizes performance as the progress made toward task completion, $\pr(\returnAtTime)=\max_t\left(\returnAtTime(t)\right)$; partial progress is included but time is still excluded.  The most common evaluation metric that considers both time and task completion is the power law discounted success rate (SPL),  $\spl(\returnAtTime,k)=\timeToReturn(1)^{-k}$; this assumes a power law relationship between time and utility.

Recent work in information retrieval has introduced preference-based measures as more sensitive, metric-free methods for evaluation \cite{moffat:freedom,diaz:rpp}.   Given two trajectories $\returnAtTimei$ and $\returnAtTimej$, we define an \textit{evaluation preference} as $\evalPreference(\returnAtTimei,\returnAtTimej)\in[-1, 1]$, where $\evalPreference > 0$ indicates that $\returnAtTimei$ is preferred, $\evalPreference < 0$ indicates $\returnAtTimej$ is preferred, and $\evalPreference = 0$ indicates indifference.  An evaluation metric can be represented as a preference by computing $\evalPreference(\returnAtTimei,\returnAtTimej)=\evalMetric(\returnAtTimei)-\evalMetric(\returnAtTimej)$.  

Although we can derive a preference from a metric, we can also directly design evaluation preferences that consider alternative systematizations.  In what follows, we will introduce several evaluation preferences that capture composite success and temporal preference systematizations and, as a result, do not require a precise relationship between measured time and success.

\begin{figure}
    \centering
    \begin{subfigure}[t]{0.3\textwidth}\centering
        \includegraphics[width=\linewidth]{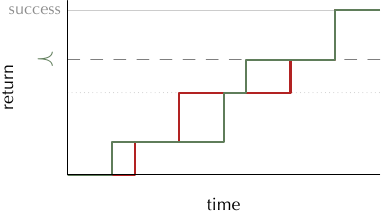}
        \caption{Lexicographic Return}\label{fig:lr}
    \end{subfigure}\hfill%
    \begin{subfigure}[t]{0.3\textwidth}\centering
        \includegraphics[width=\linewidth]{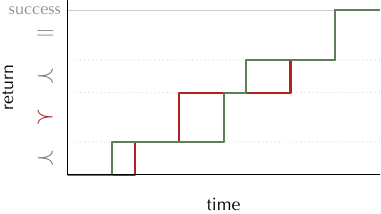}
        \caption{Return-Paired Preference}\label{fig:rpp}
    \end{subfigure}\hfill %
    \begin{subfigure}[t]{0.3\textwidth}\centering
        \includegraphics[width=\linewidth]{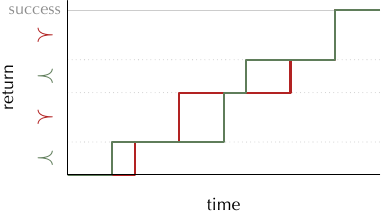}
        \caption{Interval-Paired Preference}\label{fig:ipp}
    \end{subfigure}
    \caption{\textbf{Trajectory preferences based on time to return.} Evaluation operates by comparing pairs of trajectories and aggregating preferences within trajectories.  \subref{fig:lr} Lexicographic Return ($\lr$) derives a preference from the time to reach the highest non-tied return. \subref{fig:rpp} Return-Paired Preference ($\rpp$) integrates time-to-return across all return levels. 
    \subref{fig:ipp} Interval-Paired Preference ($\ipp$) compares the time between return levels or sub-goals.
    }
\end{figure}

\paragraph{Lexicographic Return (LR).}
\label{sec:metrics:lr}
The most conservative way to design a temporal preference reproduces $\evalPreference_{\sr}$ when it is not zero and otherwise breaks ties based on time-to-return.  The lexicographic return preference (LR) follows this logic by comparing two trajectories $\returnAtTimei$ and $\returnAtTimej$  starting at the highest return level $\return = 1$ (i.e., `success').  If $\returnAtTimei$ is successful  (i.e., $\timeToReturni(\return)<\infty$) while $\returnAtTimej$ is not (i.e., $\timeToReturnj(\return)=\infty$), we say $\returnAtTimei\succ\returnAtTimej$.  If $\timeToReturni(\return)=\timeToReturnj(\return)$, we back off to the highest return level one system reaches before the other,
\begin{align}
\label{eq:lr}
    \evalPreference_{\lr} = \slSgn{\timeToReturnj(\targetReturn)-\timeToReturni(\targetReturn)}
\end{align}
where $\targetReturn = \max\{\return\in[0,1] : \timeToReturni(\return)\neq\timeToReturnj(\return)\}$.  If no such $\targetReturn$ exists, $\evalPreference_{\lr} = 0$.  Figure \ref{fig:lr} provides an example of computing LR for two trajectories. LR systematizes performance as relative temporal priority at the highest return difference. 
LR can be adopted in cases where a conservative alignment with success rate is desired, with ties broken by time or, if those are tied, lower return levels.  LR is equivalent to `lexicographic recall' in the information retrieval literature \cite{diaz:lexirecall}.

\paragraph{Return-Paired Preference (RPP).}
\label{sec:metrics:rpp}
While LR looks at a single point where two trajectories differ, return-paired preference (RPP) sweeps uniformly across all return levels and, at each level, compares the time-to-return for both trajectories.  Let $[\breakpoint_0,\ldots,\breakpoint_K]$ be the sorted union of all return levels achieved by either trajectory, augmented with $0$ and $1$.  For each return segment $[\breakpoint_{k-1}, \breakpoint_k)$, if $\timeToReturni(\breakpoint_k)<\timeToReturnj(\breakpoint_k)$, then $\returnAtTimei\succ\returnAtTimej$ for that segment.  We then average those segment-level preferences, weighted by the segment width $\breakpoint_k - \breakpoint_{k-1}$:
\begin{align}
\label{eq:rpp}
    \evalPreference_{\rpp} = \sum_{k=1}^{K} (\breakpoint_k - \breakpoint_{k-1})  \slSgn{\timeToReturnj(\breakpoint_k) - \timeToReturni(\breakpoint_k)}.
\end{align}
where  $\infty-\infty=0$.  
Figure \ref{fig:rpp} provides an example of computing RPP for two trajectories.  RPP systematizes performance as cumulative temporal advantage across all return levels, assuming a uniform weighting across levels.  As such, RPP is appropriate when we care about cumulative reward but are indifferent between return levels.  This naturally emerges in many utility gathering tasks and, as a result, the ranking analogue from information retrieval is `recall-paired preference' \cite{diaz:rpp}.

\paragraph{Interval-Paired Preference (IPP).}
\label{sec:metrics:ipp}

Rather than comparing the absolute time-to-return, the interval-paired preference compares the \emph{time increment} required to advance from one return level to the next.  This arises when rewards are accumulated as sub-goals are reached.  For each segment $[\breakpoint_{k-1}, \breakpoint_k)$, let $\delta_{\timeToReturni}^k = \timeToReturn(\breakpoint_k) -\timeToReturn(\breakpoint_{k-1})$ denote the additional time a trajectory  requires to go from return $\breakpoint_{k-1}$ to $\breakpoint_k$, and define $\delta_{\timeToReturnj}^k$ analogously.  Then,
\begin{align}
\label{eq:ipp}
    \evalPreference_{\ipp} = \sum_{k=1}^{K} (\breakpoint_k - \breakpoint_{k-1}) \cdot \slSgn{\delta_{\timeToReturnj}^k - \delta_{\timeToReturni}^k}.
\end{align}
Figure \ref{fig:ipp} provides an example of computing IPP for two trajectories.  IPP systematizes performance as local temporal efficiency at each incremental step, asking `at each sub-goal transition, which system was faster to advance?' A trajectory that starts slowly but then makes faster local progress can be preferred under IPP even if it is not preferred under RPP, because IPP compares incremental transition times whereas RPP compares absolute time-to-return. The systematized concept is closer to consistency of progress than overall speed.

We note that, when intermediate rewards are missing (e.g., only success and number of steps are recorded), the three preferences are identical, by design. 

\section{Methods and Materials}
\label{sec:methods}
We compare our parameter-free preference-based  methods---LR, RPP, and IPP---with parameter-free metric-based  methods---SR, PR, and SPL---when evaluating runs across five benchmarks, each of which contains one to six tasks, which, in turn, contain 30-500 task instances.  Our goal is to understand the relative strengths of each, with respect to meta-evaluation desiderata (Section \ref{sec:methods:evaluation}).  
Although this work focuses on parameter-free methods, we include comparisons to parameterized metrics in Appendix \ref{app:extended-baselines}.
All experiments and analysis were executed on a commodity laptop.  
\subsection{Data}
\label{sec:methods:data}
We evaluate across five benchmark families spanning interactive text, workspace, and software engineering settings.  Each benchmark dataset contains trajectories for 12-54 models, run across all task instances.  A summary of datasets can be found in Table \ref{tab:dataset-stats} of Appendix \ref{app:datasets}.  
AgentBoard (AB) \citep{ma:agentboard} provides six task suites---ALFWorld, ScienceWorld, BabyAI, PDDL, WebShop, and Tool-Query---covering heterogeneous interactive environments with partial-progress subgoal scores across 12 evaluated systems.
OpenHands-Index (OHI) \citep{openhandsindex2025} aggregates up to 22 systems across four code-generation benchmarks (SWE-bench, SWT-bench, SWE-bench-Multimodal, and GAIA).
OHI only records terminal binary success as well as the number of steps, allowing us to demonstrate the efficacy of preference-based evaluation for trajectories lacking intermediate or partial rewards. 
TheAgentCompany (TAC) \citep{agentcompany} is a suite of 175 workplace agent tasks.  While TAC records partial terminal returns and step counts, it does not include intermediate rewards. %
Text Adventure Learning Environment Suite (TALES) \citep{cui:tales} collects trajectories from roughly 50 systems across two text-adventure environments (Jericho and ScienceWorld).
We also generated the sub-goal reinforcement learning (SGRL) dataset of trajectories derived from agents acting in a variety of traditional reinforcement learning environments (DoorKey, FourRooms, Taxi) where progress can be measured by completing sub-goals.  
In addition to these datasets, we assembled two auxiliary  datasets to assess specific measure properties.  In order to test whether a measure detects a difference between models when none exists, we include a same-model variant of TALES (TALES-AA) where we use two random trajectories from the same model for each task, treating them as having come from different models; this provides null-hypothesis data for measuring false-positive rates. We use the same tasks as SGRL to define a separate `oracle' dataset (SGRL-oracle), where we have designed an optimal policy for the environment and introduced progressively more noise to provide a ground truth ordering of models (i.e., an optimal model with more interpolated noise will be inferior to a model with less noise). 
Full details of our datasets are provided in Appendix \ref{app:datasets}.

\subsection{Evaluation Measures}
\label{sec:methods:measures}
As baselines, we consider success rate (SR) as well as the (partial) terminal return (PR). We adopt a reference-free version of Success weighted by Path Length (SPL) \cite{anderson:evaluation-of-navigation-agents} by dividing the success indicator by the observed trajectory length (i.e., $k=1$).  
We measure the preference between two systems by averaging the paired preference between model outputs across the set of task instances, $\frac{1}{|\mlInputSample|}\sum_{\mlInput\in\mlInputSample}\evalPreference(\returnAtTimei_\mlInput,\returnAtTimej_\mlInput)$ where $\mlInputSample$ is the set of task instances. 
In order to generate a ranking of systems from pairwise preferences (see \cite{volkovs:generative-preference-aggregation} for a survey), we adopt the Bradley-Terry model used in ChatBot Arena \cite{chiang:chatbot-arena}.  
To do so, we extend the Bradley-Terry model to fractional labels,  mapping each preference $\evalPreference(\returnAtTimei_\mlInput,\returnAtTimej_\mlInput)$ to a soft win fraction $(\evalPreference(\returnAtTimei_\mlInput,\returnAtTimej_\mlInput) + 1)/2 \in [0, 1]$.  We then fit a standard Bradley-Terry model by maximizing the cross-entropy likelihood.  We leave the exploration of alternative aggregation methods to future work.  %
\subsection{Meta-Evaluation}
\label{sec:methods:evaluation}
In order to compare evaluation measures, we adopt several criteria from measurement theory and metric design.  

\textbf{Validity} ensures the measure actually captures the intended construct (e.g., system performance) without being redundant with existing measures of the same construct or misaligned with ground truth ordering.  
\textit{Inter-measure similarity} provides us with a data-driven understanding of the relationship between measures.  In this case, we are interested in a measure (a) being related to success rate, since both operationalize performance (i.e., convergent validity) but (b) not being so similar as to be redundant.  We measure similarity at both the instance-level (pairwise agreement between measures) as well as ranking level (Kendall's $\tau$ correlation between system rankings derived from measures).  
\textit{Oracle agreement} measures the number of true system preferences recovered by the measure; we use the SGRL-oracle dataset for these experiments.  Oracle agreement uses both sign agreement as well as the number of statistically significant preferences detected, corrected for multiple comparisons (see `Discriminative power' below).
\textit{Oracle agreement under perturbation} measures the true system preferences when we randomly drop rewards from trajectories with probability $\{0.1,0.25, 0.50\}$.

\textbf{Reliability} ensures that a measure yields stable and consistent results under resampling or small perturbations of the data.
In \textit{split-half reliability}, we randomly partition the task instances into two equal halves.  In pair analysis (P), we compute the mean metric value per model pair on each half and measure the Kendall $\pKendall$ correlation between these two lists of $O(n^2)$ model pairs. In ranking analysis (R), we compute the system ranking for each half and measure the Kendall $\pKendall$ correlation between these two lists of $n$ models.  We repeat this for 100 random splits and report the mean correlation.  Higher values indicate that the metric's system-level ranking is stable under subsampling.
In \textit{leave-one-out stability},  for each task instance, we remove it and recompute the sign of the mean difference for every model pair.  We report the fraction of pairs for which removing any single instance causes a sign flip in the aggregate preference.  A metric with few sign flips is robust to individual outlier instances.

\textbf{Sensitivity} ensures the measure can detect differences between systems.  Unlike validity measures, we are only interested in detecting a difference, not detecting an accurate difference. \textit{Tie rate} is an instance-level metric that computes the number of paired comparisons that result in a tie.   \textit{Discriminative power} computes the number of model pairs for which a metric detects a statistically significant difference \citep{sakai:evaluation-evaluation,sakai:alt-bpref,akhtar:saturation}. For each model pair and metric, we conduct a significance test using the bootstrap method; we use 10,000 replicates and obtain a two-sided p-value via the centered-statistic bootstrap. We use $\alpha=0.05$ and correct for multiple comparisons within each metric using both family-wise error rate (Holm's method) and false discovery rate (Benjamini-Hochberg). The family-wise error rate (FWER) controls the probability of making any false positive across all pairwise tests, resulting in a conservative estimate of discriminative power. By contrast, the false discovery rate (FDR) controls the expected proportion of false positives amongst the rejected hypotheses, providing a less conservative but higher-sensitivity view. 

\textbf{Data efficiency} ensures the measure achieves reliable and discriminative conclusions using as few evaluation samples as possible.  We analyze the data efficiency of a measure by observing how quickly it converges to a stable ranking of systems as a function of evaluation examples.  Specifically, we randomly downsample evaluation examples, compute the aggregated preference between system pairs, and then compute the accuracy of those preferences with respect to preferences based on the full data.  We downsample 100 times at ten points.  In addition to measuring how quickly system preferences converge, we measure how well a measure recovers the oracle system preferences as a function of number of examples, providing us with a sense of how quickly a measure converges to the correct preferences. 
\section{Results}
\label{sec:results}
Given the breadth of experiments, we include cross-benchmark aggregated results or benchmark-specific results.  Disaggregated and complete results can be found in Appendix \ref{app:validity} (Validity), \ref{app:reliability} (Reliability), \ref{app:sensitivity} (Sensitivity), and \ref{app:efficiency} (Data Efficiency).

\begin{figure}
\hfill
\begin{subfigure}[t]{0.225\textwidth}
  \centering
  \includegraphics[width=\linewidth]{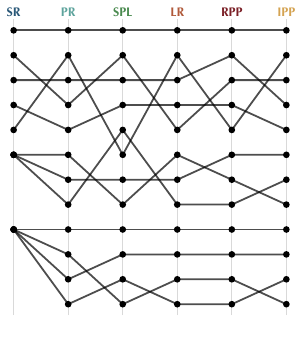}
  \caption{Bump Chart}
  \label{fig:conv-valid:bump-chart}
\end{subfigure}\hfill
\begin{subfigure}[t]{0.3\textwidth}
  \centering
  \includegraphics[width=\linewidth]{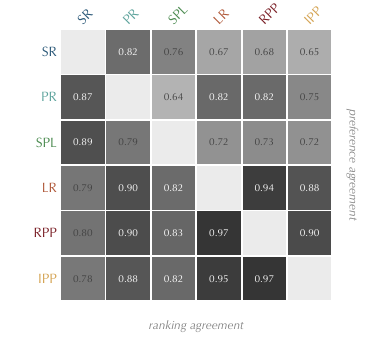}
  \caption{Inter-measure correlation}
  \label{fig:conv-valid::correlation}
\end{subfigure}\hfill
\begin{subfigure}[t]{0.225\textwidth}
  \centering
  \includegraphics[width=\linewidth]{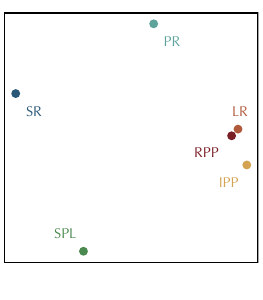}
  \caption{t-SNE embedding}
  \label{fig:mds-tau}
\end{subfigure}\hfill
\caption{
  \textbf{Inter-measure similarity:}
  \subref{fig:conv-valid:bump-chart} Bump chart showing the ranking of systems across several measures for AgentBoard ALFWorld runs; charts for other tasks can be found in Figure \ref{fig:bump-ttr}.
  \subref{fig:conv-valid::correlation} Correlation between measures.  Upper triangle: agreement in pairwise preferences.  Lower triangle: agreement in rankings.
  \subref{fig:mds-tau} t-SNE embedding of measure $\tau$ similarity. Measures close together induce similar system rankings.}
\label{fig:mds}
\end{figure}

\paragraph{Validity.}
\label{sec:results:validity}
Our results provide evidence consistent with convergent validity (high correlation with existing performance metrics) and criterion validity (improving alignment with oracle preferences) of trajectory-aware preferences. 
\textit{Inter-measure similarity.}
Figure~\ref{fig:conv-valid:bump-chart} provides an example bump chart showing how a system ranking changes across different evaluation measures. As expected, SR collapses several systems into tied ranks, which non-binary measures (PR, SPL, LR, RPP, IPP) disentangle.  While these measures generally agree on the resulting ordering, they tend to disagree on individual rank swaps. 
Figure~\ref{fig:conv-valid::correlation} scales this inter-measure similarity analysis over all of our datasets, using pairwise preference agreement (upper triangle) and Kendall's $\tau$ of system rankings (lower triangle) to compare all six measures, resulting in two clusters: a scalar success-like family (SR, PR, SPL) that compares trajectories by an aggregate scalar outcome, and a trajectory-preference family (LR, RPP, IPP) that compares them across return levels. We can visualize this clustering by using the $\tau$ correlation to construct a t-SNE embedding of measures (Figure~\ref{fig:mds-tau}).  %
\textit{Oracle agreement.}
While the convergent validity captured by our inter-measure similarity analysis provides evidence of consistency between preference-based evaluation and existing metric-based evaluation methods, agreement with oracle preferences between systems  (SGRL-oracle) provides criterion validity.  
Table~\ref{tab:summary-ttr}a reports the fraction of pairwise preferences among the degraded oracle variants that each measure recovers correctly. SR and PR both achieve only 12.8\% accuracy since their terminal-only character means they cannot distinguish successful agents whose differences appear only in how they reach the goal. SPL recovers 91.8\% of the true preferences, while the trajectory-preference measures all exceed 94\%. Requiring that oracle agreement be statistically significant demonstrates the advantage of trajectory-level preferences. Under FDR correction, RPP detects 63.2\% of true preferences as significant, followed by IPP (62.5\%) and LR (61.6\%); SPL is only able to detect 48.1\%, while SR and PR detect none. The same ordering of preferences over metrics roughly holds under the more conservative FWER correction.  %
\textit{Oracle agreement under perturbation.}
When perturbing the trajectories, we observe degradation across all measures, but preference-based measures continue to dominate SR, PR, and SPL across all perturbation levels (Figure \ref{fig:oracle-accuracy-under-perturbation}).
Additional comparisons with parameterized, full-trajectory scalar metrics (Appendix \ref{app:extended-baselines}) show that discounted return and area under the return curve achieve comparable uncorrected oracle ranking accuracy but substantially lower statistically significant recovery of correct preferences than LR, RPP, and IPP. %

\begin{figure}
\centering
\begin{subfigure}[t]{0.48\textwidth}\centering
  \includegraphics[width=\linewidth]{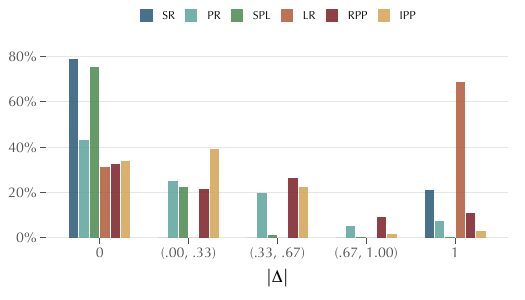}
  \caption{Measure Distribution (AB)}\label{fig:histo}\end{subfigure}\hfill
\begin{subfigure}[t]{0.48\textwidth}\centering
  \includegraphics[width=\linewidth]{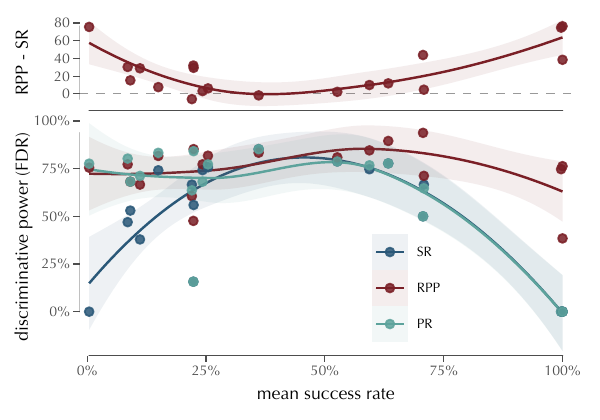} 
  \caption{FDR correction}\label{fig:sensitivity:fdr}\end{subfigure}
\caption{\textbf{Sensitivity:} (a) Distribution of measure values.  (b) Discriminative power as a function of mean task success rate.  Each point is one benchmark task.  Curves are LOWESS fits.  LR and IPP track RPP while SPL tracks SR; lines removed for clarity.}
\label{fig:sr-vs-dp}
\end{figure}

\begin{figure}
\centering
\begin{subfigure}[t]{0.48\textwidth}\centering
  \includegraphics[width=\linewidth]{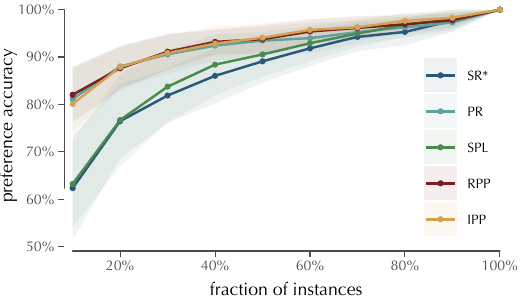}
  \caption{Ranking stability (AB/ALFWorld)}\label{fig:efficiency:stability}\end{subfigure}\hfill
\begin{subfigure}[t]{0.48\textwidth}\centering
  \includegraphics[width=\linewidth]{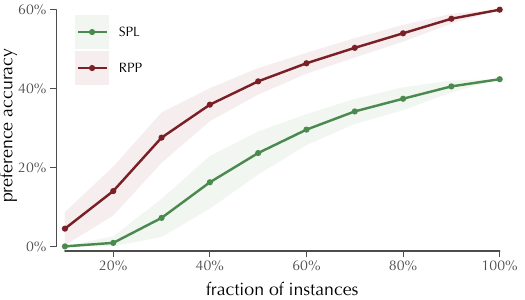}
  \caption{Oracle sig.\ accuracy (FourRooms)}\label{fig:efficiency:oracle}\end{subfigure}
\caption{\textbf{Data efficiency.}
  \subref{fig:efficiency:stability} Accuracy of model preferences based on each subsample fraction with respect to model preferences based on the full-data.  
  \subref{fig:efficiency:oracle}~Fraction of oracle pairs that are both
  correctly ordered and statistically significant (Benjamini-Hochberg correction)
  as a function of the fraction of instances used.  SR and PR omitted due to poor performance; LR and IPP omitted for clarity.}
\label{fig:data-efficiency}
\end{figure}

\paragraph{Reliability.}
\label{sec:results:reliability}
Our results demonstrate that trajectory-aware preferences produce more stable results under resampling, with fewer reversals when data is subsampled when compared with existing metric-based evaluations.
\textit{Split-half reliability.}
Table~\ref{tab:summary-ttr}b shows our results for split-half reliability.  LR and RPP both achieve high mean split-half correlations (system pair: 0.83; system ranking: 0.85), with PR also achieving correlations above 0.80. While IPP reaches a lower correlation compared to other trajectory measures, the remaining scalar measures show notably lower reliability with correlations between 0.70 and 0.75 across system pair and system ranking.  
The trajectory-preference family produces more stable rankings under subsampling than the metric family, with PR and IPP sitting between the two.
\textit{Leave-one-out stability.}
Table~\ref{tab:summary-ttr}c shows our results for leave-one-out reliability.    SR and LR both achieve a 0\% sign-flip rate: removing any single instance never reverses any pairwise system preference, primarily because removing an example will, in the worst case, turn a signed difference into a tie, not a flip.  Of measures with less discrete behavior, RPP maintains modest sign flips (1.88\%), followed by PR (2.14\%) and IPP (2.83\%). SPL shows the highest flip rate (5.5\%), consistent with its lower split-half correlation. %
\paragraph{Sensitivity.}
Our results show that, by converting ties into informative comparisons, trajectory-aware metrics recover signal that scalar measures discard, resulting in more detectable system differences.
\textit{Tie rate.}
Table~\ref{tab:summary-ttr}d shows the tie rates across measures.  As mentioned in Section \ref{sec:introduction}, SR produces ties for 74.9\% of instance comparisons, while SPL reduces ties to 63.4\% by also penalizing trajectory length for successful trajectories.  PR, on the other hand, mitigates this to 49.7\% by using terminal progress rather than binary success.
The trajectory-preference family achieves substantially lower tie rates of roughly 35\%. 
To understand tie rate, we can look at the distribution of $\evalPreference$ values for different measures.  Figure~\ref{fig:sr-vs-dp}\subref{fig:histo} shows the distribution of the absolute value of pairwise preference values at the instance level for each measure on AgentBoard (distributions for all benchmarks can be found in Appendix Figure~\ref{fig:pref-dist-all}). SR and LR concentrate nearly all of their mass at the two extremes: $|\evalPreference|=0$ (ties) and $|\evalPreference|=1$ (maximum preference), implying that individual instance comparisons carry only three values, $\{-1, 0,  1\}$. SPL, despite some mass in the middle, remains heavily bimodal. By contrast, PR, RPP, and IPP distribute mass across the range $(0,1)$, producing a richer set of preference magnitudes and lowering tie rates. When each instance comparison can take a range of values rather than only $\{-1, 0, 1\}$, the aggregate preference over many instances is more informative.
\textit{Discriminative power.}
To address concerns that lower tie rates may be due to noise, we can compute the number of statistically significant differences detected.  In general, we find that the lower tie rate in preference-based evaluation reflects greater discriminative power as demonstrated in Table~\ref{tab:summary-ttr}e.  Under FDR correction, RPP detects significant differences in 78.4\% of model pairs, with LR, IPP, and PR also above 70\%; SPL reaches 60.2\% and SR 58.5\%. Under the more conservative FWER correction, both LR and RPP provide  discriminative power above 60\%, while PR, IPP, SR, and SPL consistently detect fewer differences. These results support the higher sensitivity exhibited by preference-based evaluation, reproducing results from information retrieval research.
Figure~\ref{fig:sensitivity:fdr} shows how discriminative power varies with mean task success rate. Each point represents a single benchmark task. SR's power peaks at intermediate success rates and collapses at both extremes. When most systems either all fail or all succeed, SR cannot distinguish them. PR is able to distinguish models when they all tend to fail but, like SR, collapses as models become more successful. 
RPP maintains high discriminative power across the full range of task difficulty, including on the oracle domains where SR has zero power. %
\textit{Discriminative bias.}
We find that all measures achieve zero false-positive rates on same-model pairs (TALES-AA)  under both FWER and FDR correction (Table~\ref{tab:summary-ttr}f), providing evidence that the increase in discriminative power is not resulting in spurious differences.  %
\paragraph{Data efficiency.}
Our results show that, because instance-level comparisons are more informative, trajectory-aware measures reach stable rankings with fewer evaluation instances and converge faster to full-data conclusions. 
\textit{Ranking stability.}
Figure~\ref{fig:efficiency:stability} shows how the system preferences induced by each measure converge to the full-data model preferences as evaluation instances are subsampled.    On AgentBoard ALFWorld (shown), as well as on other benchmarks
(Appendix Figures~\ref{fig:degradation-self-ab}--\ref{fig:degradation-self-sgrl}), 
preferences based on trajectories plateau with as many or fewer 
instances when compared with preferences based on scalar metrics.  This is consistent with our reliability results, where trajectory measures demonstrate stability between downsampled datasets. 
\textit{Oracle preference recovery.}
Figure~\ref{fig:efficiency:oracle} complements this
analysis with oracle-controlled data, using a stricter criterion: a pair is
counted as correctly recovered only if the measure assigns the correct sign
\emph{and} the difference is statistically significant.  On FourRooms (shown), 
RPP's advantage is apparent at small sample sizes:
RPP's accuracy at 50--60\% of instances exceeds SPL's
full-data value.  Results on DoorKey and Taxi
(Appendix Figures \ref{fig:oracle-degradation-fwer} and \ref{fig:oracle-degradation-fdr}) exhibit the same 
pattern; the advantage is most pronounced on Taxi, the hardest domain, where
SPL detects essentially no significant pairs while RPP and the other
trajectory-preference measures still recover a non-trivial fraction.
Trajectory-level information thus improves not only oracle agreement in general but
also efficiency in detecting true system differences with statistical
confidence.

\begin{table*}
\centering
{\footnotesize
\begin{tabular}{l ccc cc c c cc cc}
\toprule
 & \multicolumn{3}{c}{(a) Oracle Rank Acc.} & \multicolumn{2}{c}{(b) Split Half} & \multicolumn{1}{c}{(c) LOO} & \multicolumn{1}{c}{(d) Tie Rate} & \multicolumn{2}{c}{(e) Disc.\ Power} & \multicolumn{2}{c}{(f) Disc. Bias} \\
\cmidrule(lr){2-4} \cmidrule(lr){5-6} \cmidrule(lr){7-7} \cmidrule(lr){8-8} \cmidrule(lr){9-10} \cmidrule(lr){11-12}
 & Acc & FW & FD & P & R &  &  & FW & FD & FW & FD \\
\midrule
SR & 12.8 & 0 & 0 & 0.75 & 0.73 & \textbf{0} & 74.9 & 44.98 & 58.47 & \textbf{0} & \textbf{0} \\
PR & 12.8 & 0 & 0 & 0.82 & 0.82 & 2.14 & 49.71 & 56.53 & 73.37 & \textbf{0} & \textbf{0} \\
SPL & 91.8 & 31.6 & 48.1 & 0.71 & 0.7 & 5.5 & 63.42 & 39.02 & 60.2 & \textbf{0} & \textbf{0} \\
\noalign{\vskip 4pt}
LR & 95.6 & 37.5 & 61.6 & \textbf{0.83} & \textbf{0.85} & \textbf{0} & \textbf{33.9} & \textbf{66.16} & 77.81 & \textbf{0} & \textbf{0} \\
RPP & 94.2 & 37.5 & \textbf{63.2} & \textbf{0.83} & \textbf{0.85} & 1.88 & 34.82 & 61.51 & \textbf{78.35} & \textbf{0} & \textbf{0} \\
IPP & \textbf{95.8} & \textbf{37.7} & 62.5 & 0.78 & 0.81 & 2.83 & 35.09 & 53.34 & 73.76 & \textbf{0} & \textbf{0} \\
\bottomrule
\end{tabular}
 }
\caption{Summary of meta-evaluation results across benchmarks. \textbf{Validity:} (a) Recovering oracle preferences.   \textbf{Reliability:} (b) split-half reliability of system pairs (P) and system rankings (R). (c) Leave-one-out sign flip rate. \textbf{Sensitivity:} (d) Instance level tie rate.  (e) Discriminative power with correction for family-wise error rate (FW) and false discovery rate (FD).  (f) Discriminative bias detection of significant differences amongst identical models.   Lower is better for LOO, tie rate, and discriminative bias; higher is better otherwise. Bold = best per column.  Details on meta-evaluation methods can be found in Section \ref{sec:methods:evaluation}.}
\label{tab:summary-ttr}
\end{table*}

\section{Discussion}
\label{sec:discussion}
Our results provide evidence that preference-based approaches can improve reliability, sensitivity, and data efficiency while preserving alignment with existing performance measures and improving agreement with oracle preferences without requiring additional data beyond trajectory logs, changes to sampling practices, reweighting, or hyperparameters.

We observe that benchmark saturation, often portrayed as the result of poor data collection or weak problem difficulty, may also be explained by the choice of evaluation measure.  Figure~\ref{fig:sensitivity:fdr} suggests that a measure like SR may be effective at distinguishing `middling' systems but fail altogether at early points in the development process (when models may be largely under-performant) or later in the development process (when models may be uniformly strong).  The intentional design of sensitive metrics allows a benchmark to more effectively  compare arbitrary populations of models.  

Notwithstanding use in information retrieval, our adoption of preference-based trajectory evaluation, while common in online or arena-style evaluation, is novel for offline evaluation outside of simple paired statistical tests.  
Existing studies in production evaluation demonstrate the effectiveness and efficiency of preference-based evaluation \cite{chapelle:interleaving}.  
Beyond this, offline preference-based win rates are comparable with arena-style win rates, allowing more consistent evaluation and avoiding any cross-metric calibration \cite{maksai:offline-online}.
At the same time, offline preference-based evaluation presents the opportunity for counterfactual preference measurement, which is impossible in online evaluation where a real user is often limited to comparing two system outputs.  

Working with temporal preference allows our measures to avoid needing to validate a precise relationship between time and utility (or return). We contrast this with discounted utility measures which, as mentioned in Section \ref{sec:motivation}, can be brittle across domains. Our methods only require that the preference be consistent across test instances without any hyperparameters.  

Finally, while we have focused on preference-based evaluation, all of our measures suggest novel methods for optimizing sequential decision-making tasks.  Avoiding the need to select a discount factor, craft partial rewards, or worry about consistent cross-task temporal discounting may allow the more efficient and robust training of models.  There is increasing evidence that preferences can be more expressive than methods that reduce performance to a scalar metric value \cite{munos2024nash,pmlr-v235-swamy24a}.  

\paragraph{Limitations} Trajectory-aware evaluation assumes that intermediate returns reflect meaningful progress toward task completion; when subgoals are poorly aligned with desired notions of utility, trajectory-based metrics are less valid. In addition, in some domains, temporal preference, while embedded in the reinforcement learning and economics literatures, may not be a desirable system property.  And while our proposed measures avoid an explicit temporal discount parameter, they
encode distinct preferences over completion, partial progress, and time.  Finally, because the strongest oracle-ranking analyses rely on synthetic environments with known optimal behavior, further work is needed to validate our findings in real-world agentic systems.  %
\section{Conclusion}
\label{sec:conclusion}

We argued that success rate as a metric discards information, compromises the efficiency of benchmarks, and leads to benchmark saturation.  By shifting to preference-based comparisons over trajectory structure, we recover evaluation signal without requiring additional data beyond trajectory logs and avoiding stronger assumptions about the relationship between time and utility. Empirically, this yields consistent gains in reliability, sensitivity, and data efficiency across benchmarks. More broadly, our results suggest that evaluation quality and benchmark utility often depend on the measurement instrument itself.     
\begin{ack}
This research was supported in part by the National Institute of Standards and Technology under Federal Award ID 60NANB24D231 and by Carnegie Mellon University's AI Measurement Science and Engineering Center (AIMSEC).     
\end{ack}

\bibliographystyle{abbrvnat} %

\appendix

\section{Use of binary metrics at ML and NLP conferences}
\label{app:conference-percentages}
We used the OpenReview API to gather abstracts for NeurIPS, NeurIPS Datasets and Benchmarks, ACL, and EMNLP between 2022 and 2025.  We then identified abstracts that contained references to any of: success rate, accuracy, exact match, task success, episode success, top-1 accuracy, pass@1, solved rate, or match rate.  Complete results are presented in Table \ref{tab:conference-percentages}.  

\begin{table*}
\centering
\small
\setlength{\tabcolsep}{8pt}
\begin{tabular}{lcccc}
\toprule
Year & ACL & EMNLP & NeurIPS Main & NeurIPS Data \\
\midrule
2022 & 8.8\%  & 8.9\%  & 13.6\% & 4.9\%  \\
2023 & 10.5\% & 10.9\% & 13.5\% & 7.8\%  \\
2024 & 10.4\% & 12.6\% & 13.0\% & 11.1\% \\
2025 & 17.0\% & 20.5\% & 18.4\% & 18.3\% \\
\bottomrule
\end{tabular}
\caption{
Percentage of published paper abstracts that reference binary metrics. 
}
\label{tab:conference-percentages}
\end{table*}

\begin{table*}
\centering
{\small
\begin{tabular}{lrrrrrrc}
\toprule
\textbf{Task} & \textbf{Models} & \textbf{Instances} & \textbf{Avg.\ length} & \textbf{Avg.\ density} & \textbf{Partial} & \textbf{Interm.} & \textbf{License} \\
\midrule
\multicolumn{7}{l}{\textbf{AgentBoard (AB)}} & GPL-2.0\\
alfworld & 12 & 134 & 2.78 & 0.1920 & \checkmark & \checkmark \\
babyai & 12 & 112 & 2.73 & 0.1336 & \checkmark & \checkmark \\
pddl & 12 & 60 & 5.55 & 0.1146 & \checkmark & \checkmark \\
scienceworld & 12 & 90 & 3.70 & 0.1340 & \checkmark & \checkmark \\
tool-query & 12 & 60 & 3.31 & 0.9720 & \checkmark & \checkmark \\
webshop & 12 & 251 & 4.58 & 0.7432 & \checkmark & \checkmark \\
\midrule
\multicolumn{7}{l}{\textbf{Openhands-Index (OHI)}} & MIT \\
gaia & 22 & 165 & 19.87 & 0.0821 & --- & --- \\
swe-bench & 21 & 500 & 67.04 & 0.0169 & --- & --- \\
swe-bench-mm & 21 & 103 & 87.94 & 0.0037 & --- & --- \\
swt-bench & 18 & 433 & 47.74 & 0.0183 & --- & --- \\
\midrule
\multicolumn{7}{l}{\textbf{TheAgentCompany (TAC)}} & N/A\\
TAC & 16 & 175 & 26.27 & 0.0380 & \checkmark & --- \\
\midrule
\multicolumn{7}{l}{\textbf{Text Adventure Learning Environment Suite (TALES)}} & N/A\\
jericho & 54 & 55 & 36.45 & 0.0980 & \checkmark & \checkmark \\
scienceworld & 53 & 30 & 32.25 & 0.2979 & \checkmark & \checkmark \\
\midrule
\multicolumn{7}{l}{\textbf{Sub-Goal Reinforcement Learning (SGRL)}} & N/A\\
doorkey & 30 & 48 & 8.23 & 0.2074 & \checkmark & \checkmark \\
fourrooms & 26 & 100 & 9.13 & 0.7198 & \checkmark & \checkmark \\
taxi & 18 & 100 & 10.71 & 0.1286 & \checkmark & \checkmark \\
\bottomrule
\end{tabular}
}
\caption{Dataset statistics. Avg.\ length: mean of max step index per trajectory. Avg.\ density: mean fraction of steps with a new return increase. Partial: any non-binary rewards present. Interm.: any trajectory has a reward between the start and final returns at an intermediate step.}
\label{tab:dataset-stats}
\end{table*}

\section{Datasets}
\label{app:datasets}
Datasets consist of agent trajectories on benchmark tasks.  
AgentBoard data \cite{ma:agentboard} downloaded from \url{https://huggingface.co/datasets/hkust-nlp/agentboard/resolve/main/data.tar.gz}.  
OpenHands Index data \cite{openhandsindex2025} downloaded on 11 April 2026 for all runs in \url{https://github.com/OpenHands/openhands-index-results}.  
TheAgentCompany data \cite{agentcompany} downloaded from \url{https://github.com/TheAgentCompany/experiments/tree/main/evaluation/1.0.0}.  
Text Adventure Learning Environment Suite data \cite{cui:tales} downloaded on 11 April 2026 from \url{https://huggingface.co/datasets/PEARLS-Lab/TALES-Trajectories}.  
To support statistical analysis, we remove tasks with fewer than 30 task instances.  

\subsection{Sub-Goal Reinforcement Learning}

The sub-goal reinforcement learning data uses three gridworld-style domains: Taxi (the Gymnasium Taxi-v3 environment), DoorKey (the MiniGrid DoorKey-5x5-v0 environment, where an agent has to pick up a key, unlock a door, and reach a goal cell), and FourRooms (the MiniGrid FourRooms environment, where four rooms are connected by single-cell hallways and an agent has to reach a goal placed in another room). For each domain   we construct a fixed set of up to 100 distinct task instances by iterating reset seeds from zero upward and retaining only the first seed whose post-reset state, characterized by a domain-specific task-defining tuple, is novel relative to all previously retained instances; the discriminating tuple is (taxi row, taxi column, passenger location, destination) for Taxi, (agent position, agent heading, key   position, door position, door-locked flag, goal position) for DoorKey, and (agent position, agent heading, goal position, goal room) for FourRooms. Each retained seed is serialized together with its decoded   state, and at evaluation time the environment is reset with the stored seed and the recorded factors are asserted to match, which keeps the set reproducible across runs. The DoorKey domain has 48 instances because we exhausted the space.

From this fixed set of tasks we generate   trajectories for a sequence of policies that span weak to strong on each domain. Two reference policies are hand-coded: a uniform random agent over the legal action set, and a deterministic oracle implemented as a   planner with full environment knowledge (a shortest-path policy over Taxi's known transition graph, and a breadth-first search over (x, y, heading) tuples that emits the relative turn/forward/pickup/toggle action    sequences needed for the two MiniGrid domains). The remaining systems are learned with standard model-free RL trained from sparse environment reward, with one seed-per-checkpoint and three (Taxi) or two   (DoorKey, FourRooms) random seeds: for Taxi we use action-masked PPO, DQN, and QRDQN over a factored one-hot symbolic observation (decomposed into taxi row, taxi column, passenger location, and destination   index), trained for 200k, 200k, and 50k steps respectively; for DoorKey and FourRooms we use PPO and A2C with a CNN policy over the default partial-observability image observation, trained for 500k and 2M steps   respectively, together with a deliberately under-budgeted ``weak'' PPO (100k for DoorKey, 500k for FourRooms, with a 32-dim feature extractor) to populate the lower end of the performance ladder. To probe the   effect of denser learning signal on the same algorithms, each learned baseline is duplicated as a ``shaped'' variant that trains the same architecture and hyperparameters under a potential-based shaping reward   computed from the symbolic state: the Taxi shaping rewards moving toward the passenger and then toward the destination, while the MiniGrid shapings reward progress toward the next subgoal in the canonical subgoal    chain (key, door, goal for DoorKey; goal-room entry then goal cell for FourRooms). At evaluation time every policy is rolled out once per  instance under a per-domain step budget (100 for Taxi, 150 for the    MiniGrid domains), and we save compact per-step logs (the decoded factored state for Taxi; agent position, carried-object type, and door-open flag for DoorKey; agent position and current room identifier for   FourRooms) so that hidden subgoal-progress labels can be recovered offline by a deterministic detector—pickup and successful drop-off for Taxi, key pickup, door opening, and goal arrival for DoorKey, and entering    the goal room and reaching the goal cell for FourRooms—producing a return-jump trace that assigns equal credit to each subgoal of a domain. 

The oracle-degradation subset keeps the same task instance set, hand-crafted oracle, and random seed fixed, and rolls out a wrapper policy that, at each step independently, replaces the oracle's chosen action with an action drawn uniformly at random from the full action space with    probability $\epsilon$; we then sweep $\epsilon$ using a two-phase calibration in which a coarse grid first brackets the value at which mean episode-return drops to roughly 80\% of oracle return, after which 19 $\epsilon$ values are chosen    linearly spaced from a small lower bound to that 80\% crossing (approximately 0.0003 to 0.0057 for Taxi, 0.010 to 0.190 for FourRooms, and 0.025 to 0.475 for DoorKey), producing for each domain a set of twenty   closely spaced systems—the oracle plus nineteen degraded variants—whose performance differences are small enough to stress the metrics under study.

\section{Validity}
\label{app:validity}
\subsection{Bump Charts}
\begin{figure*}
\centering
\begin{subfigure}[t]{0.23\textwidth}\centering
  \includegraphics[width=\linewidth]{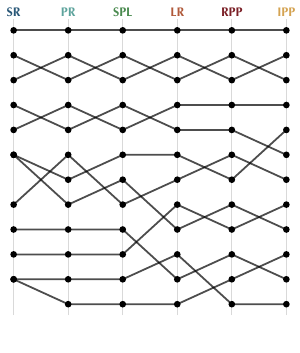}
  \caption{ALFWorld}\end{subfigure}\hfill
\begin{subfigure}[t]{0.23\textwidth}\centering
  \includegraphics[width=\linewidth]{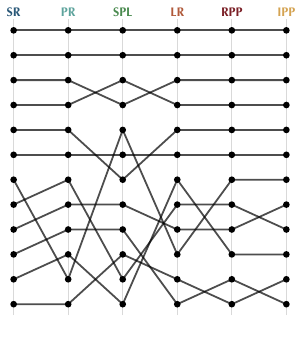}
  \caption{BabyAI}\end{subfigure}\hfill
\begin{subfigure}[t]{0.23\textwidth}\centering
  \includegraphics[width=\linewidth]{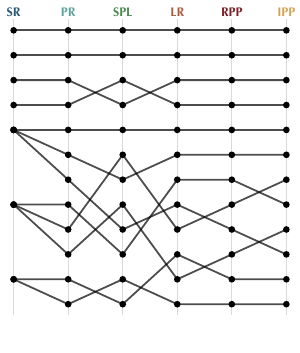}
  \caption{PDDL}\end{subfigure}\hfill
\begin{subfigure}[t]{0.23\textwidth}\centering
  \includegraphics[width=\linewidth]{graphics/scienceworld-bump.pdf}
  \caption{ScienceWorld}\end{subfigure}

\vspace{0.8em}
\begin{subfigure}[t]{0.23\textwidth}\centering
  \includegraphics[width=\linewidth]{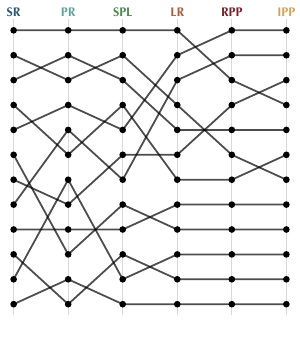}
  \caption{WebShop}\end{subfigure}\hfill
\begin{subfigure}[t]{0.23\textwidth}\centering
  \includegraphics[width=\linewidth]{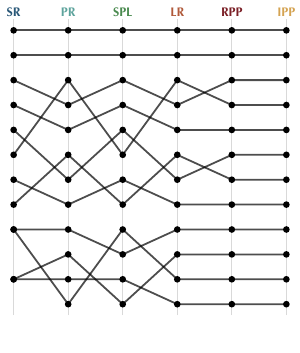}
  \caption{ToolQuery}\end{subfigure}\hfill
\begin{subfigure}[t]{0.23\textwidth}\centering
  \includegraphics[width=\linewidth]{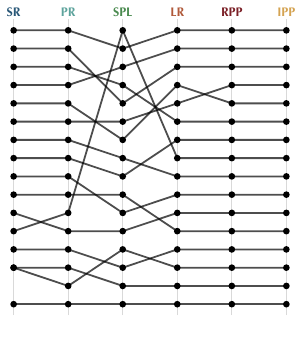}
  \caption{TAC}\end{subfigure}\hfill
\begin{subfigure}[t]{0.23\textwidth}\centering
  \includegraphics[width=\linewidth]{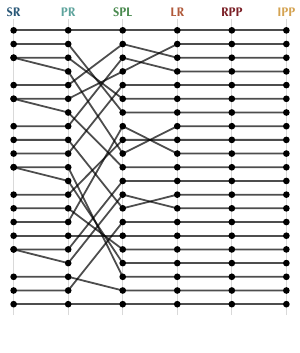}
  \caption{SWE-bench}\end{subfigure}

\vspace{0.8em}
\begin{subfigure}[t]{0.23\textwidth}\centering
  \includegraphics[width=\linewidth]{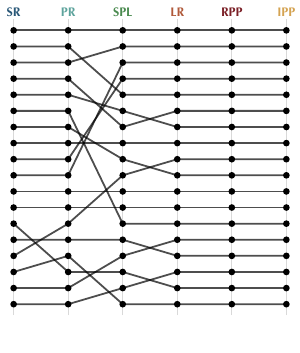}
  \caption{SWT-bench}\end{subfigure}\hfill
\begin{subfigure}[t]{0.23\textwidth}\centering
  \includegraphics[width=\linewidth]{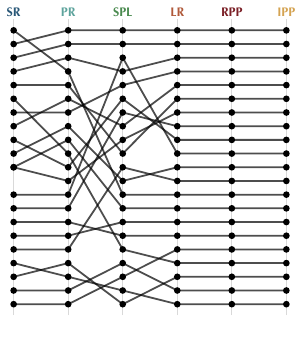}
  \caption{SWE-bench-MM}\end{subfigure}\hfill
\begin{subfigure}[t]{0.23\textwidth}\centering
  \includegraphics[width=\linewidth]{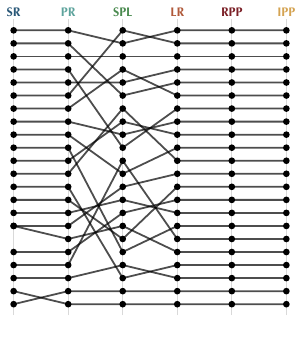}
  \caption{GAIA}\end{subfigure}%

\vspace{0.8em}
\begin{subfigure}[t]{0.23\textwidth}\centering
  \includegraphics[width=\linewidth]{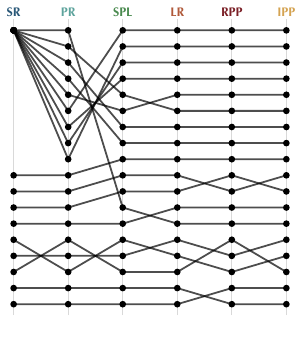}
  \caption{Taxi}\end{subfigure}\hfill
\begin{subfigure}[t]{0.23\textwidth}\centering
  \includegraphics[width=\linewidth]{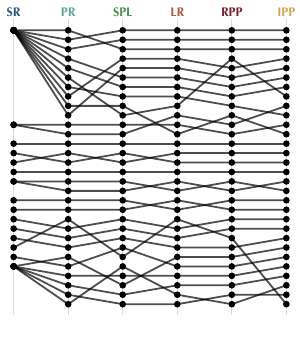}
  \caption{DoorKey}\end{subfigure}\hfill
\begin{subfigure}[t]{0.23\textwidth}\centering
  \includegraphics[width=\linewidth]{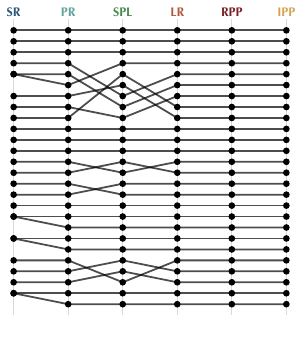}
  \caption{FourRooms}\end{subfigure}

\caption{Bump charts for tasks. Each line is one model; rank~1 is top.}
\label{fig:bump-ttr}
\end{figure*}

Figure~\ref{fig:bump-ttr} traces each model's rank across measures, one panel per task. Crossings reveal where the rankings are reordered; flat bands indicate stable rankings.

\begin{table}
\centering
{\small
\begin{tabular}{lcccc|cccc}
\hline
 & \multicolumn{4}{c|}{\textbf{Accuracy}} & \multicolumn{4}{c}{\textbf{Power (FDR)}} \\
\hline
 & Taxi & DoorKey & FourRooms & Mean & Taxi & DoorKey & FourRooms & Mean \\
\hline
SR & 0 & 0 & 38.4 & 12.8 & 0 & 0 & 0 & 0 \\
PR & 0 & 0 & 38.4 & 12.8 & 0 & 0 & 0 & 0 \\
SPL & 82.1 & 97.4 & 95.8 & 91.8 & 1.6 & 77.9 & 64.7 & 48.1 \\
\noalign{\vskip 6pt}
LR & 92.1 & 98.4 & 96.3 & 95.6 & 34.7 & 75.3 & 74.7 & 61.6 \\
RPP & 88.9 & 97.4 & 96.3 & 94.2 & 38.4 & 76.3 & 74.7 & 63.2 \\
IPP & 92.6 & 98.4 & 96.3 & 95.8 & 34.7 & 77.9 & 74.7 & 62.5 \\
\hline
\end{tabular}
 }
\caption{Oracle rank accuracy (\%) and power (\%) (C(20,2)=190 pairs per domain).}
\label{tab:oracle-rank-accuracy-top20}
\end{table}

\subsection{Oracle Rank Accuracy}
\label{sec:oracle-rank-accuracy}

To measure whether each metric predicts the correct ordering between systems
with known ground-truth performance, we construct 20 variants of an oracle
agent per domain by injecting $\varepsilon$-random actions at varying rates
($\varepsilon = 0$ for the oracle, increasing to $\approx 80\%$ of oracle
episode return at $\varepsilon_{\max}$).  This yields $\binom{20}{2}=190$
unordered pairs per domain (570 total), where the correct ordering is defined by
$\varepsilon$: a lower-noise agent is always better.

Table~\ref{tab:oracle-rank-accuracy-top20} reports rank accuracy (fraction of pairs
ranked correctly) and statistical power (fraction of correctly ranked pairs
that are also significant at $\alpha=0.05$ by two-sided bootstrap test).

\subsection{Oracle accuracy under perturbation}

\begin{figure*}
\centering
  \includegraphics[width=\linewidth]{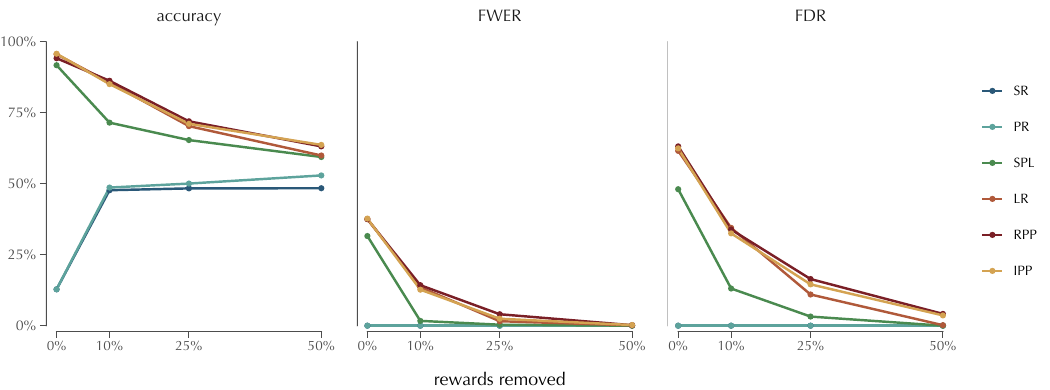}

\caption{Oracle accuracy under perturbation.}
\label{fig:oracle-accuracy-under-perturbation}
\end{figure*}

We examine the robustness of oracle preference recovery to trajectory perturbation by independently removing rewards with probabilities in \(\{0.1, 0.25, 0.5\}\). Because rewards determine the terminal return, all measures, including SR, PR, and SPL, may be affected. We evaluate each measure against the original oracle ordering, which remains unchanged by perturbation.  Figure \ref{fig:oracle-accuracy-under-perturbation} reports oracle ranking accuracy and statistically significant recovery under FWER and FDR correction. Increasing reward censoring generally reduces the accuracy of trajectory-preference measures and SPL, while SR and PR initially improve, converging to random performance of 50\%.  The improvement of SR and PR is due to ties being broken by random censoring.   LR, RPP, and IPP retain an advantage over scalar baselines. The results also suggest that aggressive reward deletion deteriorates detection of significant differences.

\label{app:oracle-accuracy-under-perturbation}
\section{Reliability}
\label{app:reliability}

Table~\ref{tab:reliability-grouped} reports split-half reliability per
metric and benchmark: instances are split into halves and we compare
both per-instance scores and the induced system rankings across the
two halves. Table~\ref{tab:sign-flip-grouped-table} complements this
with a leave-one-out stress test, giving the fraction of system pairs
whose sign of difference flips when any single instance is dropped---a
direct measure of how brittle pairwise comparisons are at each
benchmark's current size.

\begin{table}
\centering
{\footnotesize
\begin{tabular}{lcccccc|cccccc}
\hline
 & \multicolumn{6}{c|}{\textbf{System pairs}} & \multicolumn{6}{c}{\textbf{System ranking}} \\
\hline
 & AB & TAC & OHI & TALES & SGRL & Mean & AB & TAC & OHI & TALES & SGRL & Mean \\
\hline
SR & 0.82 & 0.81 & 0.66 & 0.55 & 0.92 & 0.75 & 0.79 & 0.84 & 0.68 & 0.4 & 0.93 & 0.73 \\
PR & 0.82 & 0.87 & 0.66 & 0.84 & 0.9 & 0.82 & 0.76 & 0.89 & 0.68 & 0.85 & 0.92 & 0.82 \\
SPL & 0.74 & 0.63 & 0.79 & 0.52 & 0.88 & 0.71 & 0.74 & 0.68 & 0.79 & 0.39 & 0.9 & 0.7 \\
\noalign{\vskip 6pt}
LR & 0.82 & 0.8 & 0.79 & 0.84 & 0.91 & 0.83 & 0.78 & 0.85 & 0.82 & 0.89 & 0.91 & 0.85 \\
RPP & 0.83 & 0.79 & 0.79 & 0.82 & 0.91 & 0.83 & 0.81 & 0.87 & 0.82 & 0.85 & 0.92 & 0.85 \\
IPP & 0.81 & 0.79 & 0.79 & 0.62 & 0.9 & 0.78 & 0.8 & 0.87 & 0.82 & 0.62 & 0.91 & 0.81 \\
\hline
\end{tabular}
 }

\caption{Split-half reliability, averaged within benchmark.}
\label{tab:reliability-grouped}
\end{table}

\begin{table}
\centering
{\small
\begin{tabular}{lcccccc}
\hline
 & AB & TAC & OHI & TALES & SGRL & Mean  \\
\hline
SR & 0 & 0 & 0 & 0 & 0 & 0  \\
PR & 4.55 & 0.83 & 0 & 4.89 & 0.45 & 2.14  \\
SPL & 6.57 & 10 & 5.89 & 3.03 & 2.01 & 5.5  \\
\noalign{\vskip 6pt}
LR & 0 & 0 & 0 & 0 & 0 & 0  \\
RPP & 2.53 & 0.83 & 0 & 5.21 & 0.82 & 1.88  \\
IPP & 3.54 & 0.83 & 0 & 9.32 & 0.46 & 2.83  \\
\hline
\end{tabular}

 }

\caption{Leave-one-out sign flip rate (\% of system pairs where dropping one instance changes sign of difference), averaged within benchmark.}
\label{tab:sign-flip-grouped-table}
\end{table}
\section{Sensitivity}
\label{app:sensitivity}

\subsection{Tie rate}
Table~\ref{tab:tie-rate-grouped-table} reports the fraction of
instance-level comparisons in which two system outputs receive identical
scores, averaged within each benchmark. High tie rates indicate a
metric with limited resolution---many pairs of systems are
indistinguishable on a given instance, which weakens its ability to
support fine-grained ranking.

\begin{table*}
\centering
{\small
\begin{tabular}{lcccccc}
\hline
\hline
 & AB & TAC & OHI & TALES & SGRL & Mean \\
\hline
SR & 78.88 & 80.88 & 74.56 & 84.05 & 56.1 & 74.9 \\
PR & 43.49 & 49.78 & 74.56 & 27.61 & 53.08 & 49.71 \\
SPL & 75.87 & 76.01 & 36.77 & 80.73 & 47.74 & 63.42 \\
\noalign{\vskip 6pt}
LR & 33.27 & 32.3 & 36.77 & 22.66 & 44.49 & 33.9  \\
RPP & 34.93 & 34.75 & 36.77 & 23.04 & 44.63 & 34.82  \\
IPP & 36.11 & 34.75 & 36.77 & 23.18 & 44.66 & 35.09  \\
\hline
\end{tabular}

 }
\caption{Tie rate (\% of instance comparisons with zero difference), averaged within benchmark.}
\label{tab:tie-rate-grouped-table}
\end{table*}

\subsection{Preference Distributions}
\label{sec:appendix:preference-distribution}

\begin{figure}[t]
\centering

\begin{subfigure}{0.32\linewidth}
    \centering
    \includegraphics[width=\linewidth]{graphics/AgentBoard.pdf}
    \caption{AgentBoard}
    \label{fig:pref-dist:agentboard}
\end{subfigure}
\hfill
\begin{subfigure}{0.32\linewidth}
    \centering
    \includegraphics[width=\linewidth]{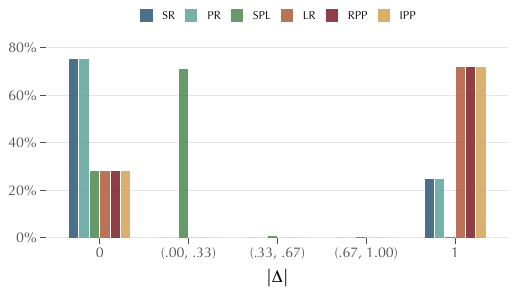}
    \caption{OpenHands}
    \label{fig:pref-dist:openhands}
\end{subfigure}
\hfill
\begin{subfigure}{0.32\linewidth}
    \centering
    \includegraphics[width=\linewidth]{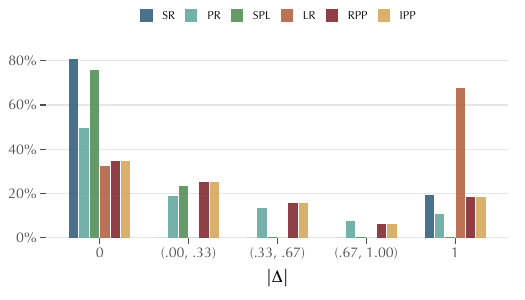}
    \caption{TheAgentCompany}
    \label{fig:pref-dist:tac}
\end{subfigure}

\vspace{0.5em}

\begin{subfigure}{0.32\linewidth}
    \centering
    \includegraphics[width=\linewidth]{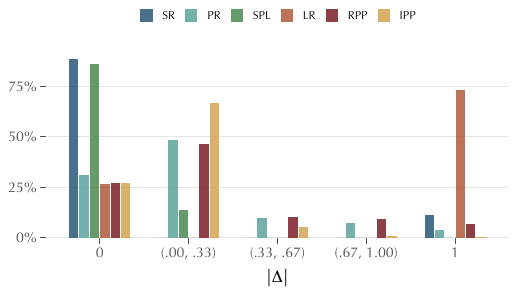}
    \caption{TALES}
    \label{fig:pref-dist:tales}
\end{subfigure}
\hfill
\begin{subfigure}{0.32\linewidth}
    \centering
    \includegraphics[width=\linewidth]{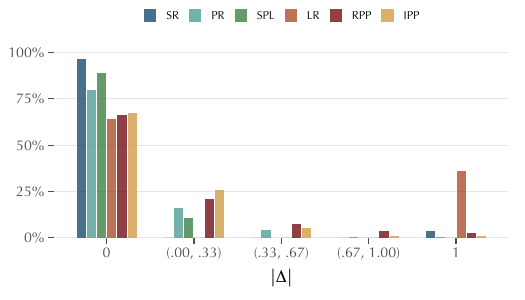}
    \caption{TALES-AA}
    \label{fig:pref-dist:tales-aa}
\end{subfigure}
\hfill
\begin{subfigure}{0.32\linewidth}
    \centering
    \includegraphics[width=\linewidth]{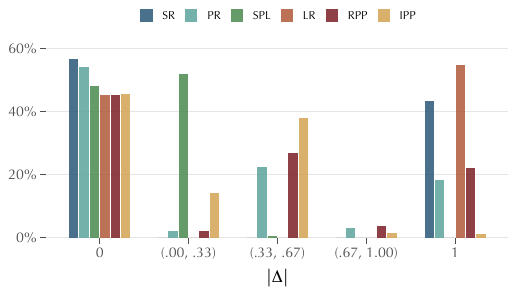}
    \caption{SGRL}
    \label{fig:pref-dist:sgrl}
\end{subfigure}

\caption{Preference distributions across all benchmark datasets.}
\label{fig:pref-dist-all}
\end{figure}

Figure~\ref{fig:pref-dist-all} shows the
distribution of absolute pairwise preference values $|\Delta|$ for each
evaluation metric, broken down by benchmark group.  Each bar shows the
proportion of instance-level comparisons whose absolute preference falls in the
indicated range.  The point masses at 0 (ties) and~1 (maximum preferences)
are shown as separate bars; the three interior bins partition $(0,1)$ into
equal thirds.

\subsection{Discriminative Power}
Table~\ref{tab:dp-grouped} reports the fraction of system pairs whose
score difference is statistically significant under a bootstrap test,
averaged within each benchmark. Higher values mean the metric resolves
more pairs of systems---a complementary view to the tie rate above,
now accounting for sampling variability rather than just exact ties.

\begin{table*}
\centering
{\footnotesize
\begin{tabular}{lcccccc|cccccc}
\hline
 & \multicolumn{6}{c|}{\textbf{FWER}} & \multicolumn{6}{c}{\textbf{FDR}} \\
\hline
 & AB & TAC & OHI & TALES & SGRL & Mean & AB & TAC & OHI & TALES & SGRL & Mean \\
\hline
SR & 47.22 & 58.33 & 38.6 & 15.97 & 64.77 & 44.98 & 59.09 & 74.17 & 55.57 & 27.98 & 75.54 & 58.47 \\
PR & 62.37 & 73.33 & 38.6 & 42.58 & 65.77 & 56.53 & 71.46 & 83.33 & 55.57 & 80.91 & 75.57 & 73.37 \\
SPL & 40.15 & 27.5 & 59.71 & 4.5 & 63.21 & 39.02 & 55.81 & 64.17 & 77.08 & 25.87 & 78.09 & 60.2 \\
\noalign{\vskip 6pt}
LR & 61.11 & 69.17 & 63.74 & 67.98 & 68.8 & 66.16 & 69.44 & 79.17 & 77.98 & 83.8 & 78.68 & 77.81 \\
RPP & 63.64 & 66.67 & 63.74 & 44.18 & 69.32 & 61.51 & 71.97 & 81.67 & 77.98 & 80.41 & 79.74 & 78.35 \\
IPP & 61.62 & 66.67 & 63.74 & 5.52 & 69.17 & 53.34 & 71.21 & 81.67 & 77.98 & 58.74 & 79.2 & 73.76 \\
\hline
\end{tabular}
 }
\caption{Discriminative power (\% of pairs with significant difference, bootstrap test), averaged within benchmark.}
\label{tab:dp-grouped}
\end{table*}

\section{Data efficiency}
\label{app:efficiency}

\subsection{Stability}
\label{sec:degradation-self}

\begin{figure*}
\centering
\begin{subfigure}[t]{0.30\textwidth}\centering
  \includegraphics[width=\linewidth]{graphics/alfworld.pdf}
  \caption{ALFWorld}\end{subfigure}\hfill
\begin{subfigure}[t]{0.30\textwidth}\centering
  \includegraphics[width=\linewidth]{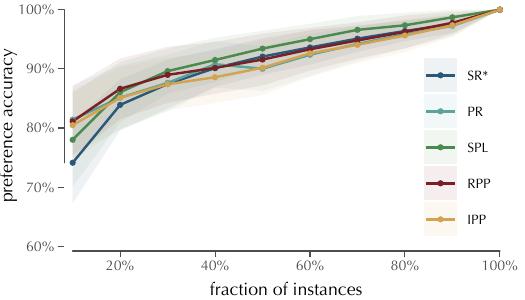}
  \caption{BabyAI}\end{subfigure}\hfill
\begin{subfigure}[t]{0.30\textwidth}\centering
  \includegraphics[width=\linewidth]{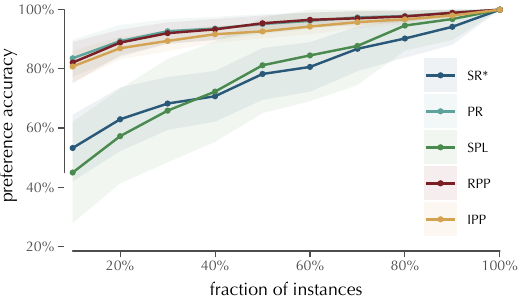}
  \caption{PDDL}\end{subfigure}

\vspace{0.8em}
\begin{subfigure}[t]{0.30\textwidth}\centering
  \includegraphics[width=\linewidth]{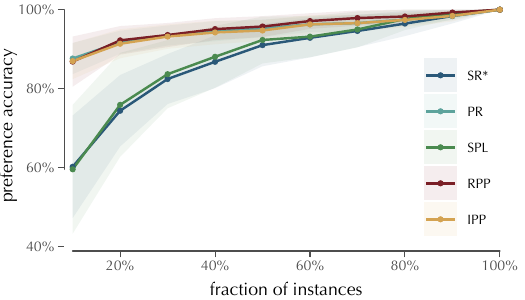}
  \caption{ScienceWorld}\end{subfigure}\hfill
\begin{subfigure}[t]{0.30\textwidth}\centering
  \includegraphics[width=\linewidth]{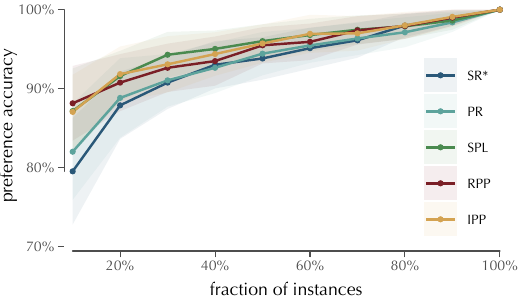}
  \caption{WebShop}\end{subfigure}\hfill
\begin{subfigure}[t]{0.30\textwidth}\centering
  \includegraphics[width=\linewidth]{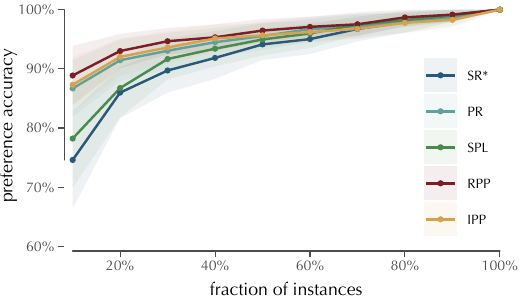}
  \caption{ToolQuery}\end{subfigure}

\caption{Preference preservation (self-reference) --- AgentBoard tasks.}
\label{fig:degradation-self-ab}
\end{figure*}

\begin{figure*}
\centering
\begin{subfigure}[t]{0.30\textwidth}\centering
  \includegraphics[width=\linewidth]{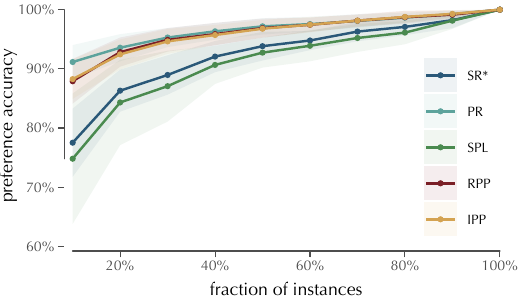}
  \caption{TAC}\end{subfigure}\hfill
\begin{subfigure}[t]{0.30\textwidth}\centering
  \includegraphics[width=\linewidth]{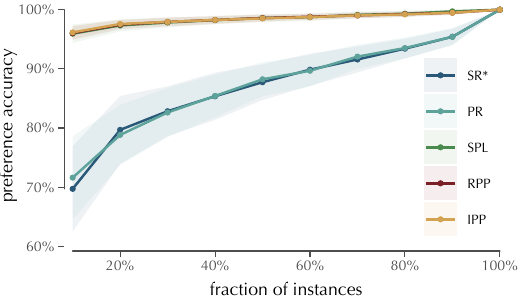}
  \caption{SWE-bench}\end{subfigure}\hfill
\begin{subfigure}[t]{0.30\textwidth}\centering
  \includegraphics[width=\linewidth]{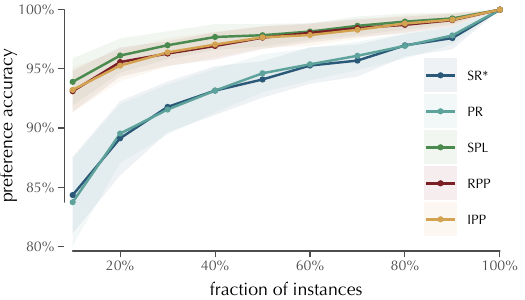}
  \caption{SWT-bench}\end{subfigure}

\vspace{0.8em}
\begin{subfigure}[t]{0.30\textwidth}\centering
  \includegraphics[width=\linewidth]{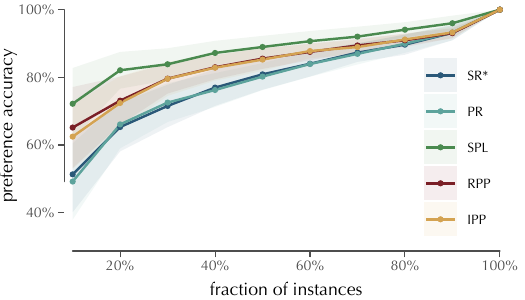}
  \caption{SWE-bench-MM}\end{subfigure}\hfill
\begin{subfigure}[t]{0.30\textwidth}\centering
  \includegraphics[width=\linewidth]{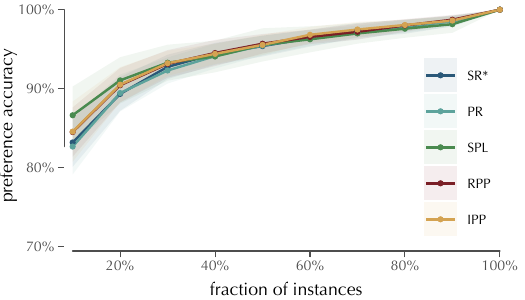}
  \caption{GAIA}\end{subfigure}

\caption{Preference preservation (self-reference) --- TAC and OHI tasks.}
\label{fig:degradation-self-tac-ohi}
\end{figure*}

\begin{figure*}
\centering
\begin{subfigure}[t]{0.30\textwidth}\centering
  \includegraphics[width=\linewidth]{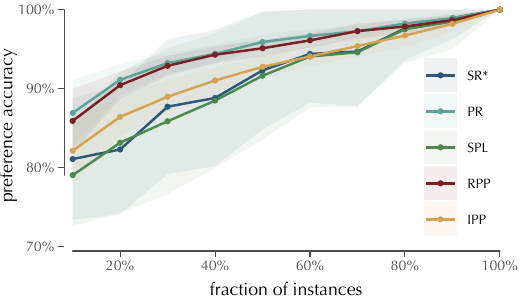}
  \caption{Jericho}\end{subfigure}\hfill
\begin{subfigure}[t]{0.30\textwidth}\centering
  \includegraphics[width=\linewidth]{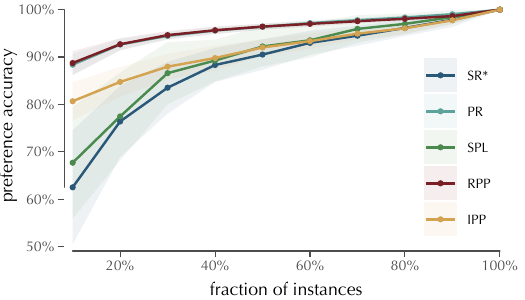}
  \caption{ScienceWorld}\end{subfigure}

\caption{Preference preservation (self-reference) --- TALES tasks ($\geq 30$ instances).}
\label{fig:degradation-self-tales}
\end{figure*}

\begin{figure*}
\centering
\begin{subfigure}[t]{0.30\textwidth}\centering
  \includegraphics[width=\linewidth]{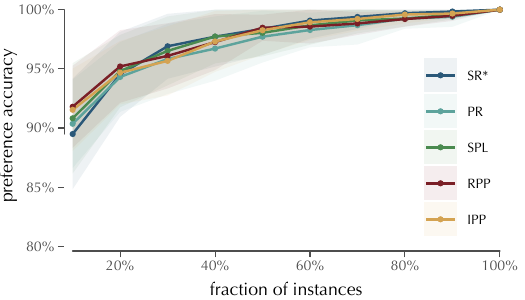}
  \caption{Taxi}\end{subfigure}\hfill
\begin{subfigure}[t]{0.30\textwidth}\centering
  \includegraphics[width=\linewidth]{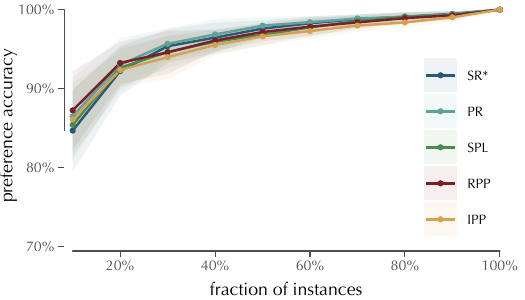}
  \caption{DoorKey}\end{subfigure}\hfill
\begin{subfigure}[t]{0.30\textwidth}\centering
  \includegraphics[width=\linewidth]{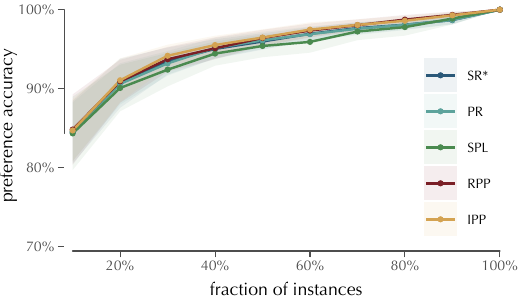}
  \caption{FourRooms}\end{subfigure}

\caption{Preference preservation (self-reference) --- SGRL tasks.}
\label{fig:degradation-self-sgrl}
\end{figure*}

Figures~\ref{fig:degradation-self-ab}--\ref{fig:degradation-self-sgrl}
show how each metric's pairwise preferences degrade as the evaluation
budget shrinks. For each sample fraction, we measure how often the
preference computed on the subset agrees with the preference computed
on the full instance set (self-reference). Curves that stay near 1
indicate a measure whose system preferences are stable under aggressive subsampling; curves that fall off quickly suggest the metric needs the full benchmark to be trustworthy.

\subsection{Oracle recovery}
\begin{figure}
\centering
\begin{subfigure}[t]{0.32\textwidth}\centering
  \includegraphics[width=\linewidth]{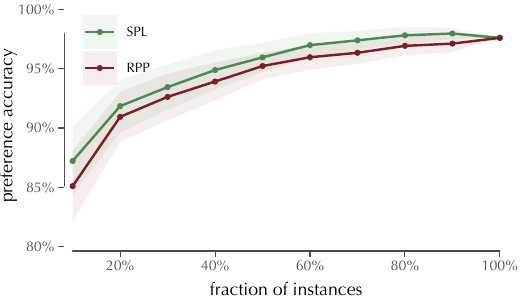}
  \caption{DoorKey}\end{subfigure}\hfill
\begin{subfigure}[t]{0.32\textwidth}\centering
  \includegraphics[width=\linewidth]{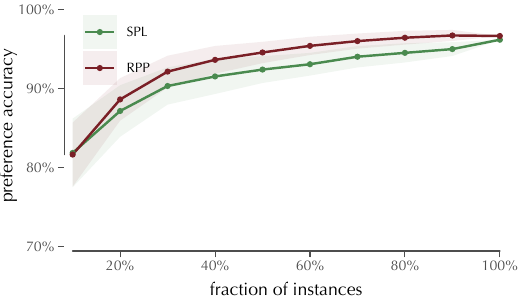}
  \caption{FourRooms}\end{subfigure}\hfill
\begin{subfigure}[t]{0.32\textwidth}\centering
  \includegraphics[width=\linewidth]{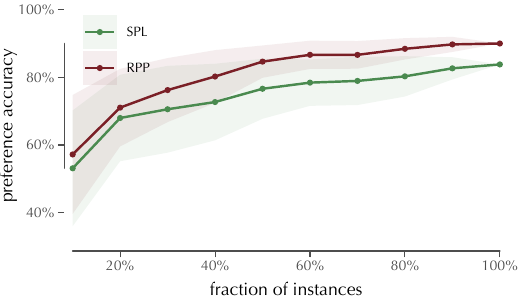}
  \caption{Taxi}\end{subfigure}
\caption{Oracle sign accuracy as a function of the fraction of instances used to compute each metric.  Pairs are drawn from the degraded oracle variants.  RPP recovers the correct pairwise preference more reliably than SPL at all sample sizes, with the advantage most visible in the more difficult Taxi domain.}
\label{fig:oracle-degradation}
\end{figure}

\begin{figure}
\centering
\begin{subfigure}[t]{0.32\textwidth}\centering
  \includegraphics[width=\linewidth]{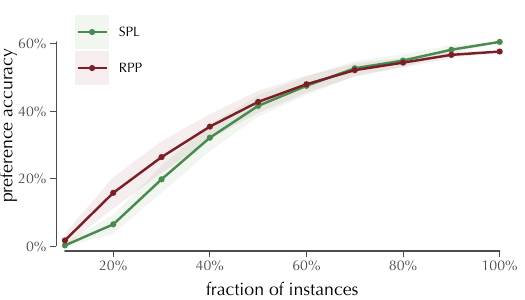}
  \caption{DoorKey}\end{subfigure}\hfill
\begin{subfigure}[t]{0.32\textwidth}\centering
  \includegraphics[width=\linewidth]{graphics/oracle_degradation_fourrooms_sig_fwer.pdf}
  \caption{FourRooms}\end{subfigure}\hfill
\begin{subfigure}[t]{0.32\textwidth}\centering
  \includegraphics[width=\linewidth]{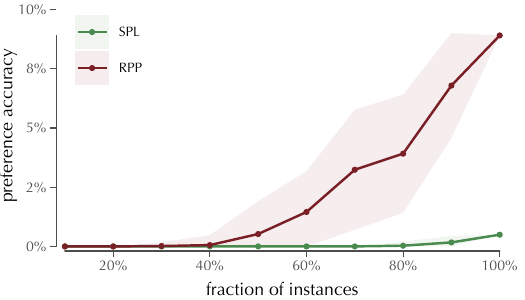}
  \caption{Taxi}\end{subfigure}
\caption{Oracle significant accuracy (FWER, Holm correction) as a function of the fraction of instances.  The fraction of oracle pairs that are both correctly ordered \emph{and} statistically significant after family-wise error rate correction.}
\label{fig:oracle-degradation-fwer}
\end{figure}

\begin{figure}
\centering
\begin{subfigure}[t]{0.32\textwidth}\centering
  \includegraphics[width=\linewidth]{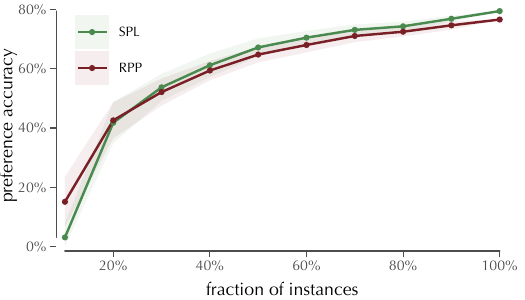}
  \caption{DoorKey}\end{subfigure}\hfill
\begin{subfigure}[t]{0.32\textwidth}\centering
  \includegraphics[width=\linewidth]{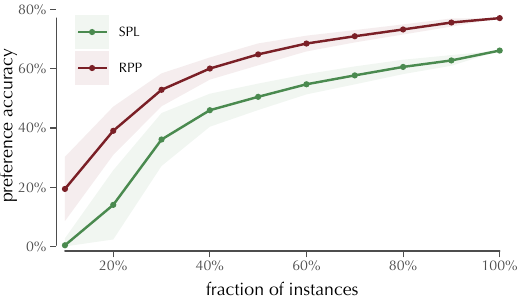}
  \caption{FourRooms}\end{subfigure}\hfill
\begin{subfigure}[t]{0.32\textwidth}\centering
  \includegraphics[width=\linewidth]{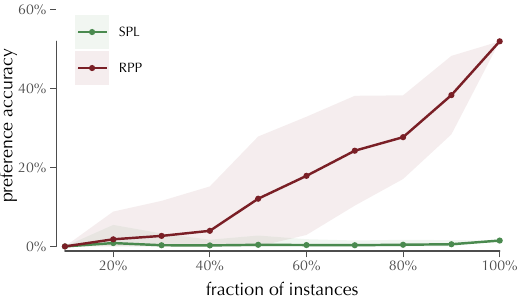}
  \caption{Taxi}\end{subfigure}
\caption{Oracle significant accuracy (FDR, Benjamini-Hochberg correction) as a function of the fraction of instances.  The fraction of  oracle pairs that are both correctly ordered \emph{and} statistically significant after false discovery rate correction.}
\label{fig:oracle-degradation-fdr}
\end{figure}

Figures~\ref{fig:oracle-degradation}--\ref{fig:oracle-degradation-fdr}
show how each measure recovers the ground-truth ordering
between degraded oracle variants as the fraction of
instances used shrinks. Figure~\ref{fig:oracle-degradation} reports
raw sign accuracy (fraction of oracle pairs ordered correctly), and
Figures~\ref{fig:oracle-degradation-fwer}
and~\ref{fig:oracle-degradation-fdr} report the stricter joint
criterion of correct ordering \emph{and} statistical significance
under FWER (Holm) and FDR (Benjamini-Hochberg) correction.

\section{Comparison with parameterized metrics}
\label{app:extended-baselines}
We consider two parameterized full-trajectory scalar baselines: discounted return (DR), with discount factor \(\gamma=0.99\), and area under the return curve (AUC), computed over a common horizon set to the maximum trajectory length in the benchmark. These baselines help compare the efficacy of parameter-free preference-based evaluation with parameterized metrics. Table \ref{tab:extended-baselines} presents these extended results.  
\begin{table*}
\centering
{\footnotesize
\begin{tabular}{l ccc cc c c cc cc}
\toprule
 & \multicolumn{3}{c}{(a) Oracle Rank Acc.} & \multicolumn{2}{c}{(b) Split Half} & \multicolumn{1}{c}{(c) LOO} & \multicolumn{1}{c}{(d) Tie Rate} & \multicolumn{2}{c}{(e) Disc.\ Power} & \multicolumn{2}{c}{(f) Disc. Bias} \\
\cmidrule(lr){2-4} \cmidrule(lr){5-6} \cmidrule(lr){7-7} \cmidrule(lr){8-8} \cmidrule(lr){9-10} \cmidrule(lr){11-12}
 & Acc & FW & FD & P & R &  &  & FW & FD & FW & FD \\
\midrule
SR & 12.8 & 0 & 0 & 0.75 & 0.73 & \textbf{0} & 74.9 & 44.98 & 58.47 & \textbf{0} & \textbf{0} \\
PR & 12.8 & 0 & 0 & 0.82 & 0.82 & 2.14 & 49.71 & 56.53 & 73.37 & \textbf{0} & \textbf{0} \\
SPL & 91.8 & 31.8 & 48.1 & 0.71 & 0.7 & 5.5 & 63.42 & 39.02 & 60.2 & \textbf{0} & \textbf{0} \\
\noalign{\vskip 4pt}
AUC & 94.2 & 28.6 & 47.2 & \textbf{0.83} & 0.83 & 3.18 & 34.06 & 59.54 & 76.27 & \textbf{0} & \textbf{0} \\
DR & 94 & 27.7 & 47.2 & \textbf{0.83} & 0.84 & 2.11 & \textbf{33.9} & \textbf{67.16} & \textbf{78.65} & \textbf{0} & \textbf{0} \\
\noalign{\vskip 4pt}
LR & 95.6 & 37 & 61.4 & \textbf{0.83} & \textbf{0.85} & \textbf{0} & \textbf{33.9} & 66.16 & 77.81 & \textbf{0} & \textbf{0} \\
RPP & 94.2 & 37.4 & \textbf{63.7} & \textbf{0.83} & \textbf{0.85} & 1.88 & 34.82 & 61.51 & 78.35 & \textbf{0} & \textbf{0} \\
IPP & \textbf{95.8} & \textbf{37.9} & 62.5 & 0.78 & 0.81 & 2.83 & 35.09 & 53.34 & 73.76 & \textbf{0} & \textbf{0} \\
\bottomrule
\end{tabular}
 }
\caption{Summary of meta-evaluation results across benchmarks. \textbf{Validity:} (a) Recovering oracle preferences.   \textbf{Reliability:} (b) split-half reliability of system pairs (P) and system rankings (R). (c) Leave-one-out sign flip rate. \textbf{Sensitivity:} (d) Instance level tie rate.  (e) Discriminative power with correction for family-wise error rate (FW) and false discovery rate (FD).  (f) Discriminative bias detection of significant differences amongst identical models.   Lower is better for LOO, tie rate, and discriminative bias; higher is better otherwise. Bold = best per column.  Differences with Table \ref{tab:summary-ttr} are due to randomness in the analysis.  Details on meta-evaluation methods can be found in Section \ref{sec:methods:evaluation}.}
\label{tab:extended-baselines}
\end{table*}

\end{document}